\pdfoutput=1
\documentclass{article}
\usepackage[numbers]{natbib}

     \usepackage[preprint]{neurips_2019}

\usepackage[utf8]{inputenc} 
\usepackage[T1]{fontenc}    
\usepackage{hyperref}       
\usepackage{url}            
\usepackage{booktabs}       
\usepackage{amsfonts}       
\usepackage{nicefrac}       
\usepackage{microtype}      
\usepackage{graphicx}
\usepackage{cleveref}
\usepackage{caption} 
\usepackage{multirow}
\usepackage[normalem]{ulem}
\useunder{\uline}{\ul}{}
\usepackage[normalem]{ulem}
\useunder{\uline}{\ul}{}

\title{Accelerated Cloud for Artificial Intelligence (ACAI)}

\author{%
  Dachi CHEN \\
  Language Technologies Institute\\
  Carnegie Mellon University\\
  Pittsburgh, PA 15213 \\
  \texttt{dachic@andrew.cmu.edu} \\
  \And
  Weitian DING \\
  Language Technologies Institute\\
  Carnegie Mellon University\\
  Pittsburgh, PA 15213 \\
  \texttt{weitiand@andrew.cmu.edu} \\
  \And
  Chen LIANG \\
  Language Technologies Institute\\
  Carnegie Mellon University\\
  Pittsburgh, PA 15213 \\
  \texttt{chenlia2@andrew.cmu.edu} \\
  \And
  Chang XU \\
  Language Technologies Institute\\
  Carnegie Mellon University\\
  Pittsburgh, PA 15213 \\
  \texttt{changx@andrew.cmu.edu} \\
  \And
  Junwei ZHANG \\
  Language Technologies Institute\\
  Carnegie Mellon University\\
  Pittsburgh, PA 15213 \\
  \texttt{junwei@andrew.cmu.edu} \\
  \And
  Majd SAKR \\
  Computer Science Department\\
  Carnegie Mellon University\\
  Pittsburgh, PA 15213 \\
  \texttt{msakr@cs.cmu.edu} \\
}

\begin{document}
\maketitle
\begin{abstract}
    Training an effective Machine learning (ML) model is an iterative process that requires effort in multiple dimensions. Vertically, a single pipeline typically includes an initial ETL (Extract, Transform, Load) of raw datasets, a model training stage, and an evaluation stage where the practitioners obtain statistics of the model performance. Horizontally, many such pipelines may be required to find the best model within a search space of model configurations. Many practitioners resort to maintaining logs manually and writing simple glue code to automate the workflow. However, carrying out this process on the cloud is not a trivial task in terms of resource provisioning, data management, and bookkeeping of job histories to make sure the results are reproducible.

    We propose an end-to-end cloud-based machine learning platform, Accelerated Cloud for AI (ACAI), to help improve the productivity of ML practitioners. ACAI achieves this goal by enabling cloud-based storage of indexed, labeled, and searchable data, as well as automatic resource provisioning, job scheduling, and experiment tracking. Specifically, ACAI provides practitioners (1) a data lake for storing versioned datasets and their corresponding metadata, and (2) an execution engine for executing  ML jobs on the cloud with automatic resource provisioning (auto-provision), logging and provenance tracking. To evaluate ACAI, we test the efficacy of our auto-provisioner on the MNIST handwritten digit classification task, and we study the usability of our system using experiments and interviews. We show that our auto-provisioner produces a 1.7x speed-up and 39\% cost reduction, and our system reduces experiment time for ML scientists by 20\% on typical ML use cases.  

\end{abstract}

\section{Introduction}
    Research in AI has long been a sophisticated task that requires practitioners to have expertise in both systems and machine learning. Although the key part of an AI task is modeling, practitioners spend a significant amount of time dealing with repetitive work like data management, resource orchestration, and bookkeeping. This work is time-consuming, and members within the same team may have their own methods of exploring the search space, making a machine learning project hard to maintain over time and upon team member turnover. Machine learning practitioners can benefit from an end-to-end cloud-based platform that manages the entire life cycle of the machine learning model development process and allows them to focus on developing and tuning their models instead of worrying about data and job management, or cloud resource orchestrations.
    
    We defined the scope of the ACAI system as being able to resolve the following user pain points:
    \begin{enumerate}
      \item \textbf{Data management.} Practitioners often maintain their own data repository, and thus introduces overhead to manage and share data among team members. The problem exacerbates during iterative ML processes, which produce a significant amount of intermediate data. It becomes difficult even for the data owner to find the input and output data corresponding to a certain job after many iterations of experiments. ACAI should provide storage solutions that facilitate data sharing with proper version control and search mechanisms.
      \item \textbf{Resource provisioning and orchestration.} When using the cloud, practitioners need to learn how to provision resources and move data around on the cloud. It is also important to properly estimate the resource requirements of jobs in order to perform data processing or model training tasks within the desired time and monetary budget, which requires knowledge in cloud instance types and pricing models. ACAI should be able to prevent users from the chores of interacting with cloud services. As long as the resource requirements are provided, the job execution is fully managed. Furthermore, ACAI can proactively try to find the best resource configurations for a job to save user’s time or cost.
      \item \textbf{Data and Model provenance.} Practitioners perform iterations of experiments in producing an effective model. This typically generates numerous intermediate datasets, models, and experimental results that are poorly managed without context information or may even be lost. This limits the ability to reflect on and reproduce historical results. ACAI helps users track data and model provenance with proper context information and reproduce past experiment results.
    \end{enumerate}
    
    We propose ACAI to address these pain points. ACAI consists of a data lake and an execution engine. The data lake manages ML artifacts and keeps track of experiments using a directed acyclic graph where nodes are datasets and edges are ML jobs. The execution engine provisions cloud resources and executes ML jobs with an auto-provisioning feature that recommends optimal resource configurations for launching an ML job. 
    
    In the following sections, we first present related work. Next, in the design section, we discuss the key system abstractions, experiment tracking, and the auto-provisioning algorithm. In the implementation details section, we present a micro-service oriented architecture that implements the ACAI system. In the experiment section, we compare the job runtime and cost of auto-provisioned resource configuration to a baseline using the MNIST handwritten digit recognition task as an example. We obtain a 1.7x speedup and 39\% cost saving respectively by fixing cost and optimizing for runtime, and fixing runtime and optimizing for cost. We also conduct a usability study where the control runs a hyperparameter tuning task using the Google Cloud Platform (GCP), and the treatment performs the same task using ACAI. We show that the treatment achieves comparable accuracy in the task by using 20\% less time and 2\% less cost. Finally, we demonstrated our system and interviewed an ML practitioner who provided feedback on early iterations of our system. The interview with the ML practitioner shows that our system reduces the overhead in organizing experiment results, and provides appealing features for ML researcher such as interactive provenance tracking.

\section{Related Work}
    A typical Machine Learning (ML) workflow requires a data lake solution for storing, accessing, and sharing of datasets, features and modeling artifacts, an execution platform for resource provisioning and scheduling ML jobs, and a provenance server that tracks experiments. In this section, we compare existing frameworks and services along these dimensions. 
    
    \subsection{Data Lake}
    Different from conventional data warehouses, a data lake provides a data storage solution that is better suited for machine learning tasks. Key features of a data lake include the storage of heterogeneous datasets along with their metadata, centralized access points, on-demand access, and namespaces on datasets \cite{brackenbury2018draining, terrizzano2015data, hai2016constance, chessell2014governing, quix2016gemms,  gupta2009answering, gao2018navigating, walker2015personal, rangarajan2015scalable, yu2023inkgan, wang2023sentiment}. Various data lakes may have different focuses. For example, \cite{brackenbury2018draining} proposed an approach that utilizes multi-dimensional similarity of files to facilitate data discovery and management but does not have complete solutions for collecting those information such as data provenance. GEMMS\cite{quix2016gemms} tries to extract and manage various metadata from heterogeneous data, but as to the accessibility, such as metadata querying, is still not very mature.

    Some variations of data lakes also support queries over datasets, schema extraction, and dashboards that display data insights \cite{hai2016constance, chessell2014governing, gao2018navigating, hu2020antvis}. Microsoft’s Azure Data Lake (ADL), for example, consists of two main parts: Data Lake Store and Data Lake Analytics. Data Lake Store, backed by Azure Blob Storage, provides a Hadoop-compatible file system, hierarchical namespaces, and fine-grained access control lists. However, aiming to serve general applications, ADL does not support machine learning use cases where data objects are usually annotated with rich metadata for experiment tracking and reproducibility. It also does not provide metadata-based file searching or file versioning solutions. AWS Lake Formation offers another data lake solution. Powered by many other AWS services, it supports data importing from various sources and low-cost high-reliability backup. But similar to ADL, it does not provide metadata-based searching and experiment tracking for ML use cases. 
    
    \subsection{Execution Engine}
    
    Mainstream cloud providers have built execution platforms for ML on top of their suite of cloud services, but some gaps exist. Azure Machine Learning\cite{mund2015microsoft} (Azure ML) enables automated machine learning and machine learning lifecycle management on Azure. However, low-level Azure ML services, such as Blob Storage and Linux Virtual Machine, require significant effort in orchestration to serve common data engineering and machine learning use cases, while high-level services, such as Machine Learning Service and Face API, do not support experimenting with customized models. Amazon SageMaker\cite{joshi2020amazon}, a fully managed cloud-service, features an automated annotation workflow and efficient distributed training with AWS-optimized Tensorflow. It also supports serving a model in production. While Amazon SageMaker utilizes many AWS features such as auto-scaling, GPU computing, and distributed training, it does not support the profiling of ML jobs and optimizing job resource utilizations. SageMaker is better suited for deploying and scaling a working ML workflow instead of development and experimentation. 

    But the cloud-based execution engines are resource consuming, while resource provisioning requires domain knowledge. To address that, Jalaparti et al. \cite{jalaparti2012bridging} from Microsoft research purposed the Bazaar framework that allows users to specify the performance goals of MapReduce jobs and the framework predicts the resources needed by profiling the runtime on a sample dataset supplied by the client. While Bazaar optimizes resource pool utilization for MapReduce jobs, it does not apply to single-node computing libraries such as scikit-learn and Tensorflow.
    
    \subsection{End-to-end Machine Learning Systems}
    
    With the development of data lake technologies and cloud-based execution platforms, recent years have seen developments in frameworks for managing feature engineering and machine learning workflows. TensorFlow Extended (TFX)\cite{baylor2017tfx}, built on top of Google's TensorFlow\cite{abadi2016tensorflow}, is an end-to-end production-oriented machine learning pipeline. It enables "continuous training" and "production-level scalability". Since Tensorflow supports model serialization, trained models can be seamlessly handed off to deployment and serving. However, TFX does not manage provenance features such as model versioning and experiment tracking. Schelter, et, al.\cite{schelter2017automatically} proposed a lightweight system for tracking metadata and provenance of ML experiments. The system strictly requires lineage information of models, which allows data scientists to make model comparisons and reproduce models with the same dataset and code. However, they introduced client-side libraries built on top of selected APIs of ML frameworks (SparkML, MXNet, scikit-learn, etc.) that monitors data transformations and model parameters, making the system framework-specific and requires users to incorporate new code into their ML programs for automated metadata extraction. Zaharia et, al.\cite{zaharia2018accelerating} proposed MLFlow, which addresses experiment tracking, reproducibility, and model packaging and deployment. Unlike the framework in \cite{schelter2017automatically}, MLFlow defines an open API for common ML use cases, and thus it supports any ML tools as long as the users implement MLFlow’s API. But the API has to be implemented in Python. Also, while MLFlow provides integration with cloud services for model serving, it does not have a cloud-based execution platform for feature engineering and model training.

    We have seen many end-to-end machine learning frameworks, and some of them are cloud-based. However, no mainstream systems provide both provenance tracing and an execution engine with automatically resource provisioning and profiling. ACAI is designed to fill in this gap to further facilitate the workflow of machine learning practitioners.

\section{System Design}
\begin{figure}[!htb]
        \begin{center}
        \includegraphics[scale=0.2]{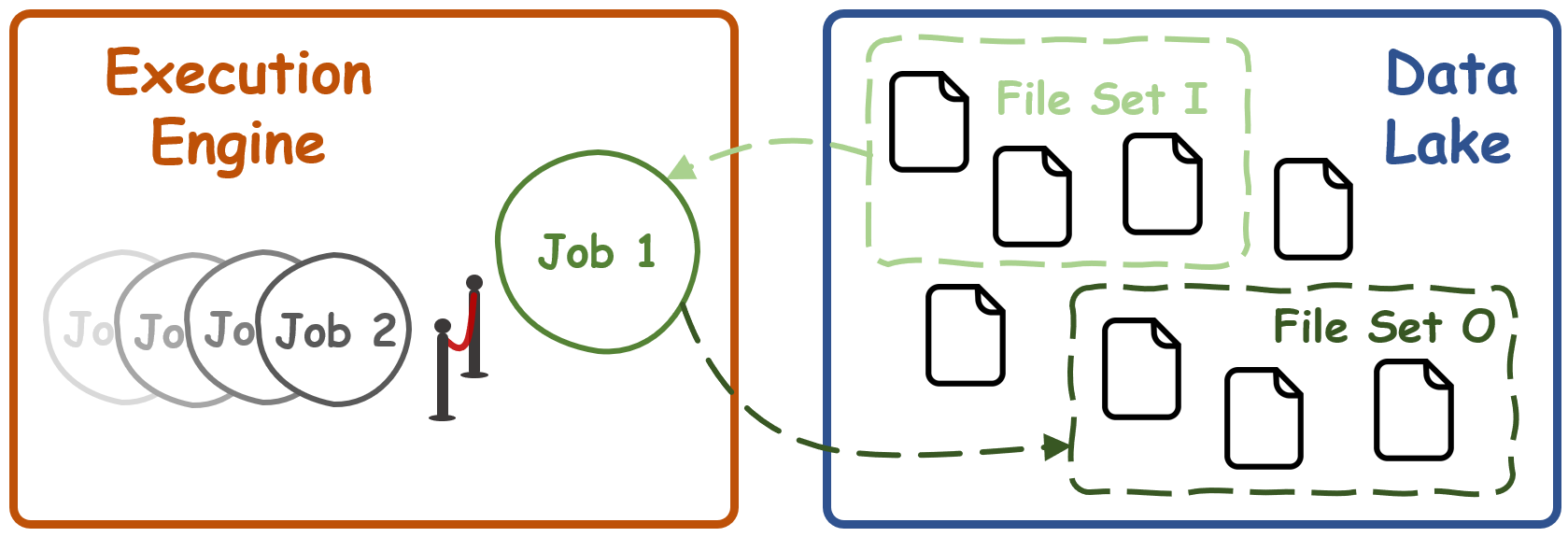}  
        \caption{ACAI System Abstractions}
        \label{fig:abs}
        \end{center}
    \end{figure}

    In a typical ML project, scientists make incremental changes to features and ML models and evaluate the changes against a testing dataset. They usually keep track of the experiments by journaling the experiments or building leaderboards of successful models.  ACAI provides the following abstractions to capture the ML activities, 
    \begin{enumerate}
    \item \textbf{Project} is an isolated workspace with data, jobs, and users.
    \item \textbf{User} is an individual client that interacts with the platform. 
    \item \textbf{File Set} is a set of version-specific cloud-stored files that can be created with low storage and computation overhead. 
    \item \textbf{Job} is an encapsulation of an ML program, including its input and output file set, runtime environment, code, and arguments. 
    \end{enumerate}
    We designed a data lake and an execution engine to model the machine learning activities, which include feature engineering, model training, and evaluations (\autoref{fig:abs}). Users interact with the ACAI system via a command-line tool and a dashboard.
    \begin{enumerate}
    \item \textbf{Data Lake} stores files, the files’ metadata, and the provenance graph. It contains a versioned file system and supports retrieval of file metadata and provenance.
    \item \textbf{Execution Engine} is a platform for the execution of ML jobs. It manages the submission, execution, monitoring, and persistence of ML jobs. It also provides resource auto-provisioning features that automatically decide the optimal resource configuration for an ML job based on the job characteristics.
    \end{enumerate}
    In this section, we discuss the design of the ACAI system by introducing access control, the entry point of the system, followed by the details of data lake. We then present the design of the execution engine and discuss metadata and provenance stores. Finally, we show the ACAI user interface.
    
    \subsection{Access Control}
    ACAI uses the concept of \textbf{User} and \textbf{Project} to manage various user data and resources. Projects are isolated workspaces with several member users. All the member users have access to the data under the project.  ACAI adopts a low-overhead, token-based authentication procedure. Each user has a unique token that can be used to send requests to ACAI or to log in to the web dashboard. Furthermore, each project has an administrator user who is authorized to create new users under the project, while a global administrator user of the entire deployed platform is authorized to create new projects.
    
    \subsection{Data Lake}
    The ACAI data lake has an S3 backed data storage that provides Posix-like semantics for accessing files. A file set glues a list of files in data lake. Both files and file set has metadata for experiment tracking. The dependency between file sets and jobs is presented using the provenance graph. In this section, we discuss the major components of data lake in detail. 
        \subsubsection{Data Storage}
        ACAI provides cloud data storage where data is stored hierarchically as files. It supports upload, download, and list. Every user-submitted job runs against data in the data storage and stores its output data back to the data storage.

        Files in ACAI data storage support \textbf{versioning}. Files can be overwritten by simply uploading new versions, while data lake persists all historical versions, and the provenance graph always tracks specific file versions. The latest versions are used if a version number is not explicitly specified, and older versions can be accessed by suffixing the file name with a specific version number, such as /data/train.json\:2. 
        
        \subsubsection{File Set}
        The concept of file set is built on top of the versioned cloud storage and is designed to represent the input or output of job executions. They are light-weight entities that only stores a list of references to versioned files. File set provides a higher-level abstraction for job executions and enables a larger granularity in provenance tracking than what individual files provide. 

        File set also supports versioning.  Clients create a new version of a file set by specifying a list of files from scratch. Similar to versioned files, a file set can be specified with or without an explicit version number, and by default, the latest version is used.
        
        Every job takes a file set as input. Versioned files in the data set are downloaded to the job container’s local file system as unversioned files. Therefore,  a file set cannot contain multiple versions of the same file. As a result, clients can specify a file version using a file set version. For example, "/data/train.json@HotpotQA:1" specifies the file /data/train.json with the version that is referenced by the file set "HotpotQA:1". File set version can also be used to "filter out" files in the underlying file storage hierarchy. For example,  "/data/@HotpotQA:1" refers to all files under the directory "/data/" that are part of the file set "HotpotQA:1".
        
        The introduction of file set appears to require extra effort from the user to explicitly create and manage file sets.  We show how data lake helps to reduce the bookkeeping overhead. First, since file set is an abstraction built on top of the file system, it takes no extra effort when the user is only interacting with the underlying file system, for example, listing directory contents. File sets can be built voluntarily when the user needs to run jobs. Second, due to the flexibility of file specification, data lake provides convenience function to create, update, merge and create subsets of file sets, in which data lake automatically builds dependencies between the source and target file sets. Here are some examples:
        \begin{enumerate}
            \item Merging:\\
            \texttt{create\_file\_set(‘MergedQA’,[‘/@HotpotQA’,‘/@ColdpotQA’])}\\
            Create a new file set called MergeQA that merges all the files from two file sets: HotpotQA and ColdpotQA. Dependency is built from MergedQADataSet to HotpotQA and ColdpotQA.
            
            \item Updating:\\
            \texttt{create\_file\_set(‘HotpotQA’,[‘/@HotpotQA’,‘/data/train.json’])}\\
            Update the file set HotpotQA by keeping all the content of the latest version and adding a new file named ‘/data/train.json’ (or update the existing /data/train.json). Dependency is built from the new-version HotpotQA to the old-version HotpotQA.
            
            \item Subsetting:\\
            \texttt{create\_file\_set(‘HotpotQAValidationSet’,[‘/validation/@HotpotQA’])}\\
            Create a new file set called HotpotQAValidationSet that contains only files under the validation directory of file set HotpotQA. Dependency is from HotpotQAValidationSet to HotpotQA.
        \end{enumerate}
        
        In the design process, we considered an alternative to file set, which is to support versioning on folders and use folders as the input and output of jobs. We call this method “versioned folder”. Versioned folder does not require users to manage new abstractions, and also supports file copy with no extra storage usage. However, it can be unnecessarily complicated when versioned folders are nested with versioned folders and files, which brings difficulty in implementation and can be quite tricky for users to maintain. Also, it requires all job inputs to be copied into a single directory.

        \subsubsection{Metadata}
        Metadata of files, file sets, and jobs are key-value attributes. ACAI supports retrieval of files, file sets, and jobs through matching key-value attributes. These attributes include built-in metadata such as creation time that is extracted at file upload, file set creation, and job execution. In addition, ACAI allows users to supply custom metadata in the form of tags or key-value pairs through CLI and the dashboard. ACAI also provides an intelligent log parser that parses user logs and attaches metadata to file sets or experiments automatically at job runtime. To trigger the automatic tagging in the log parser, user can print log in a special format as follows:
        \begin{center} \texttt{[ACAI\_TAG/ACAI\_TAG\_NUM] FILESET  KEY:VALUE} \end{center}
        Users can retrieve artifacts through the ACAI CLI by providing key-value pairs, and ACAI supports range queries such as time range queries as well as max/min queries. An exemplar query could be to find all the file sets that are generated by jobs submitted by John(creator) today (create\_time) that use model BERT (model) with a result precision that is higher than 0.5 (precision).
        
        \subsubsection{Provenance}
\begin{figure}[!htb]
            \begin{center}
            \includegraphics[scale=0.4]{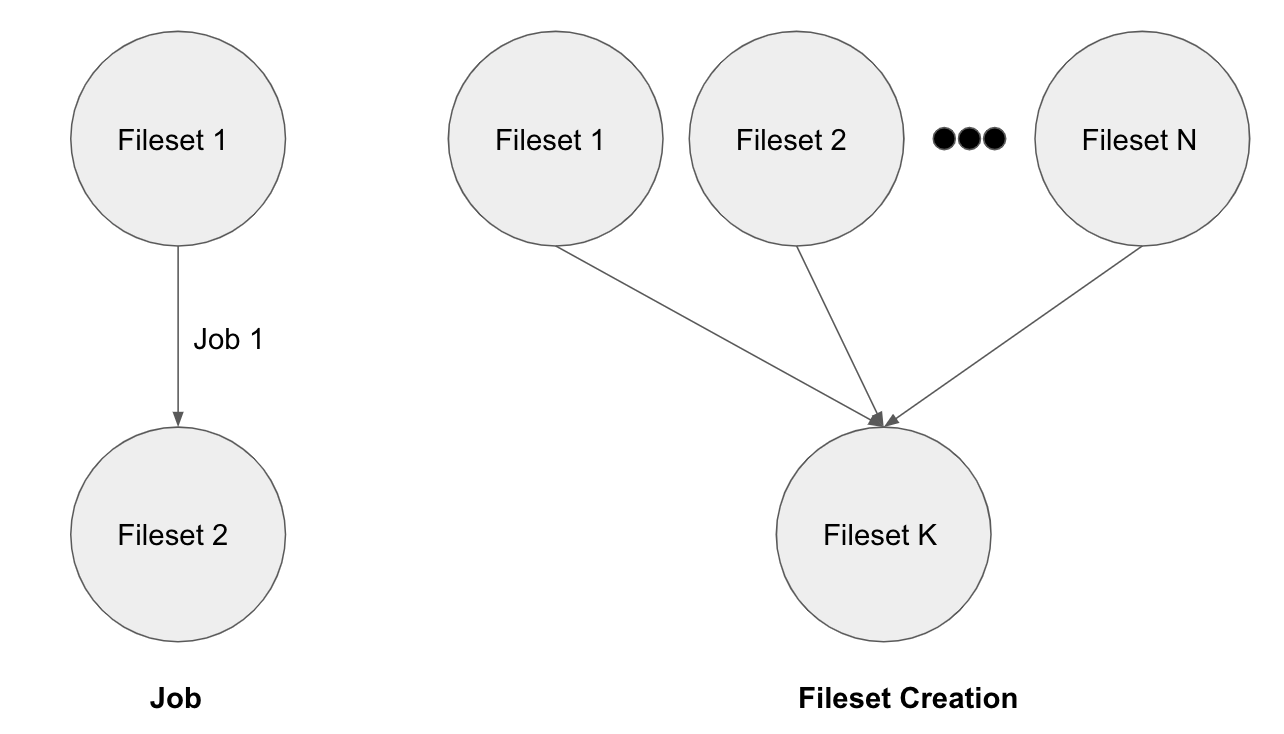}  
            \caption{Relationship between two nodes}
            \label{fig:pro_graph}
            \end{center}
        \end{figure}

        Provenance information of datasets and models is important to the machine learning community in that:
        \begin{enumerate}
            \item It enforces the reproducibility of training pipelines.
            \item It helps scientists keep track of experiment iterations and facilitates the analysis of experiments.
        \end{enumerate}
        ACAI keeps track of all the actions performed inside the system and supports the retrieval of such provenance information. As illustrated in \autoref{fig:pro_graph}, the provenance graph of the ACAI system is a directed acyclic graph built by file sets and actions performed in the system. Specifically, each file set inside ACAI is a node, and an action is an edge. There are two types of actions, file set creations, and job executions.
        
    \subsection{Execution engine}
    The execution engine is stateless and has two responsibilities, job scheduling, and resource provisioning. The scheduler manages the life cycles of jobs. It maintains $n$ job queues, one for each (project, user) tuple. Jobs are scheduled in a FIFO manner. Resource provisioning can be supervised or automatic. In the supervised setting, users supply resource configurations. In the automatic setting, the execution engine searches for the optimal resource configuration. We call this process auto-provisioning. Auto-provisioning can be broken down into two subproblems: a supervised learning problem, and a constrained optimization problem. 
    
        \subsubsection{Job Life Cycle}
\begin{figure}[!htb]
            \begin{center}
            \includegraphics[scale=0.2]{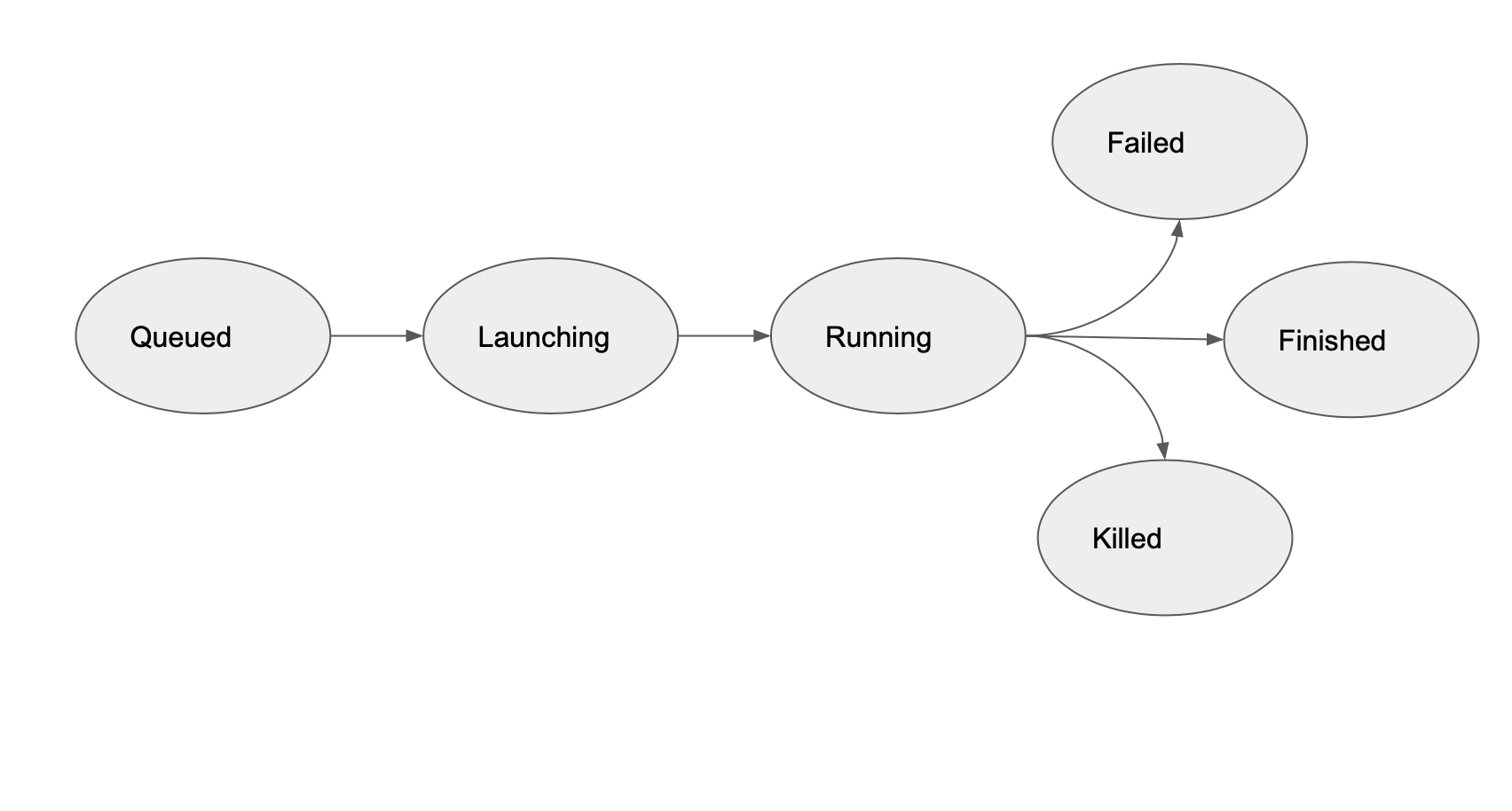}  
            \caption{Job Life Cycle}
            \label{fig:job_life}
            \end{center}
        \end{figure}

        In ACAI, the (input file set, job, output file set) triplet is immutable, so a job can only be submitted and scheduled once. A job belongs to a (project, user) tuple, and each (project, user) tuple has a FIFO queue of jobs. The life cycle of a job can be represented using a state machine (\autoref{fig:job_life}). A (project, user) tuple can have a maximum of $k$ jobs in the launching and running state. This policy contributes to the fairness of resource allocation in that the system cannot be overflowed by jobs from a single user. Once the job is submitted, it immediately enters the FIFO queue. When the number of launching or running state is less than k, the job enters the launching state where the execution engine provisions a container for the job execution. The execution engine has a buffer of the launching jobs. A job enters the running state once the resource requirement of the job can be satisfied. A running job can either finish or fail, depending on the return code of the user program. Job can be killed by the user at any time.
        
        \subsubsection{Resource Auto-provisioning}
        Auto-provisioning is the process of finding the optimal resource configuration of an ML job.  We categorize auto-provisioning into two tasks: (1) optimize job cost by fixing the maximum job runtime and (2) optimize job runtime by fixing the maximum cost. We approach auto-provisioning by first learning to predict the runtime of an ML job conditioned on resource configuration and command-line arguments. Predicting the cost of a job is trivial given a predicted runtime since we model cost as $g = \mu_c c f + \mu_m m f$, where $c$ is the number of  CPU cores, and $m$ is memory, $\mu_c$ is the unit cost of CPU and $\mu_m$ is the unit cost of memory.

        Suppose an ML job contains $k$ configurable command-line arguments such as the number of epochs, $\tau_1,\cdots,\tau_k$, then we learn a function, $f$, to predict the runtime of the job $\hat{t} = f(\tau_1,\cdots,\tau_k,c,m)$, where $c$ is the number of CPU cores and $m$ is memory allocated in megabytes (MB). This can be cast as a supervised learning problem where we obtain a training set by exploring a subset of the full configuration space $(\tau_1,\cdots,\tau_k,c,m)$. Once $f$ is learned, auto-provisioning becomes a constrained optimization problem. In the case of optimizing runtime, we minimize $f$ subject to $g \leq C$, where $C$ is the maximum cost. Optimize cost fixing the runtime is similar.
        
    \subsection{User Interface}
    ACAI system offers a web-based dashboard, a command-line interface (CLI), and a Python software development kit (SDK) for programmatic access. 

\begin{figure}[!htb]
            \begin{center}
            \includegraphics[scale=0.2]{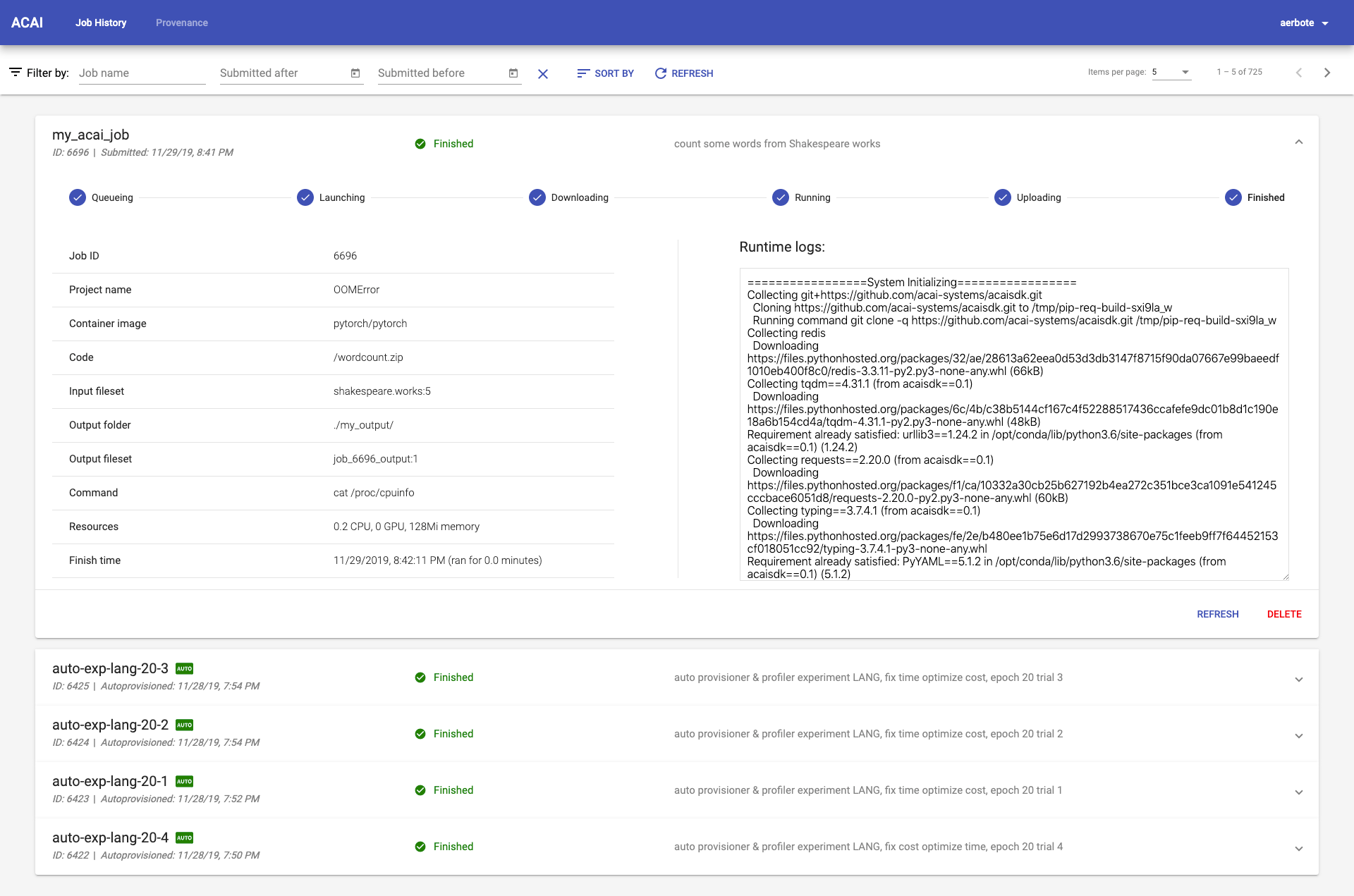}  
            \caption{Dashboard job history page}
            \label{fig:dash_job_page}
            \end{center}
    \end{figure}

\begin{figure}[!htb]
            \begin{center}
            \includegraphics[scale=0.26]{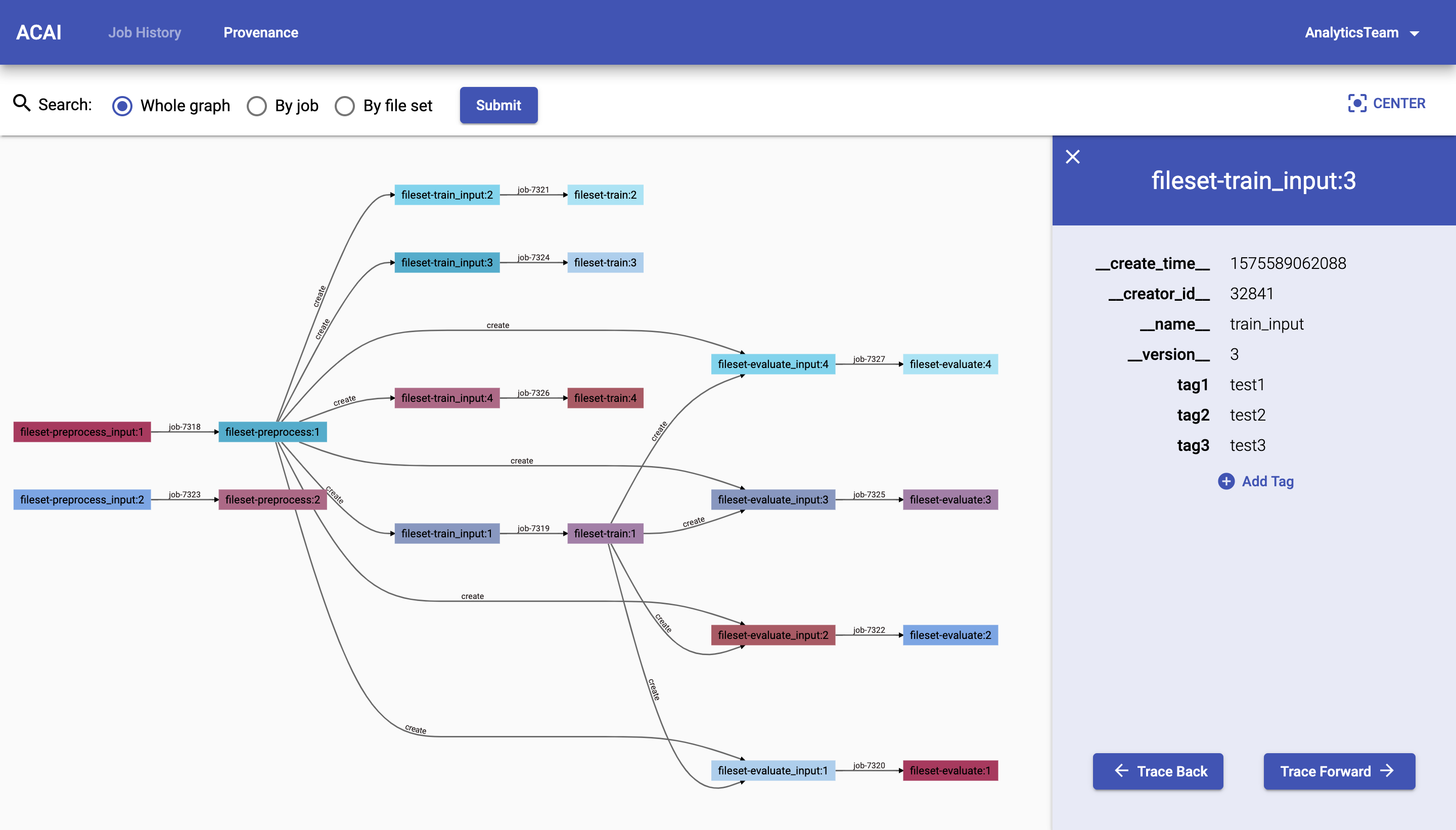}  
            \caption{Dashboard provenance page}
            \label{fig:dash_pro_page}
            \end{center}
    \end{figure}

    ACAI dashboard is a web application that enables users to monitor the progress of submitted jobs and keep track of provenance. The dashboard contains two main pages. The job history page (\autoref{fig:dash_job_page}) shows all jobs submitted by the user with detailed information, including current status, metadata, and runtime logs. During job execution, status and logs are updated on the dashboard in real-time. This page also provides additional functionalities such as job filtering, sorting, and pagination. The provenance page (\autoref{fig:dash_pro_page}) displays a provenance graph that illustrates the relationships of all the jobs and file sets within the project. Besides showing the entire graph, we also support interactive provenance tracing, which enables users to trace back or forward from any single file set. Finally, users can examine the full list of metadata tagged for each file set and job on the same page.
    
    For every service available to the user, corresponding SDK API and CLI command are provided. Documentation can be found here: \url{https://acai-systems.github.io/acaisdk/}.
    
\section{Implementation Details}
\begin{figure}[!htb]
        \begin{center}
        \includegraphics[scale=0.55]{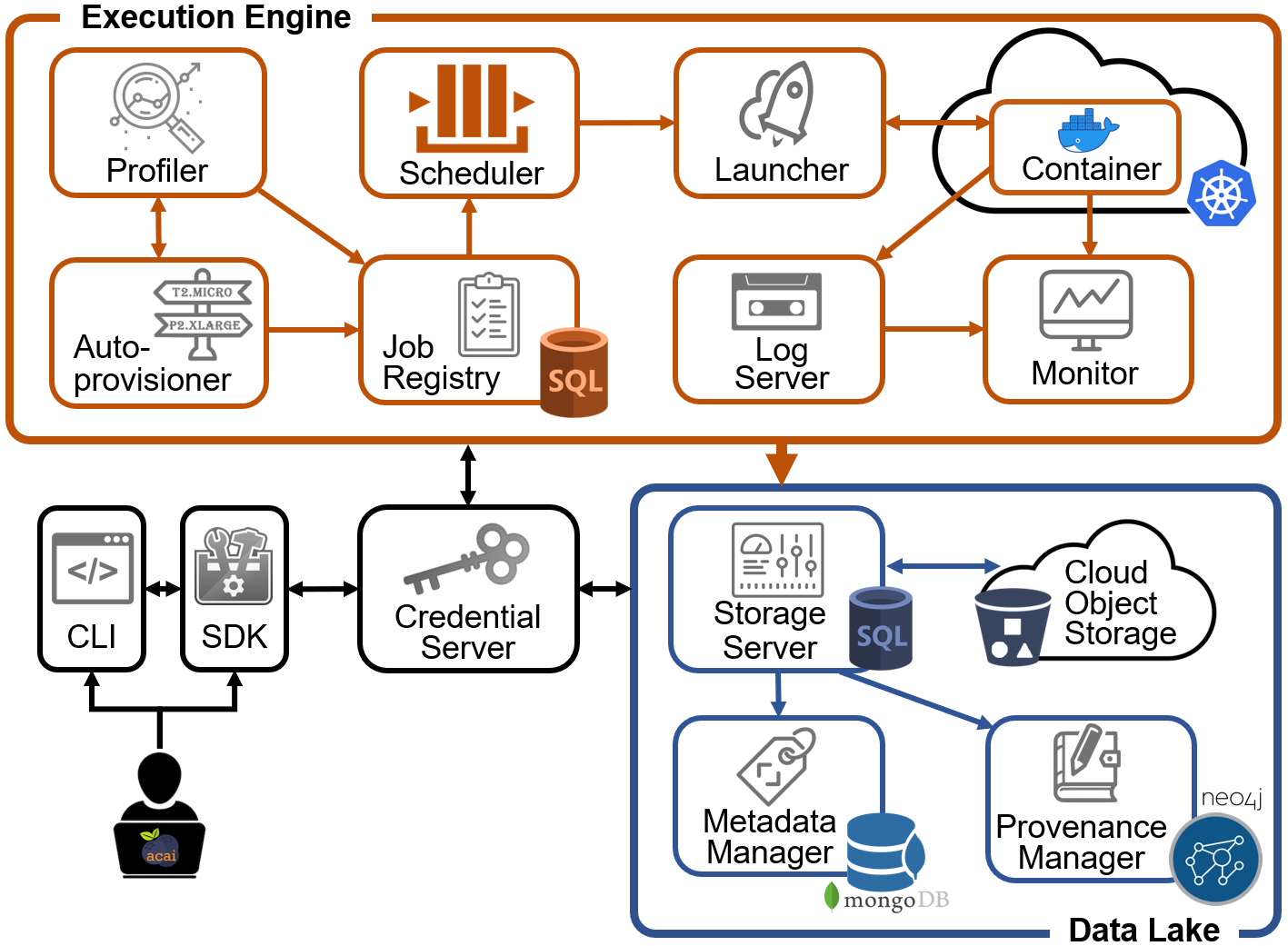}  
        \caption{ACAI system architecture}
        \label{fig:acai-arch}
        \end{center}
    \end{figure}

As illustrated in \autoref{fig:acai-arch}, the implementation of ACAI adopts a microservices oriented architecture. Credential Server is the entry point to our system that provides access control. Execution engine, contains four core microservices, a job scheduler, a job launcher, a job monitor, and a job registry. They together manage the submission, execution, monitoring, and persistence of ML jobs. The execution engine also includes a profiler and an auto-provisioner. Data lake consists of the storage server, the metadata manager, and the provenance manager. In this section, we first introduce the implementation of the credential server. We then present the details of job execution flow and resources auto-provisioning by the microservices of the execution engine. We also introduce how the storage server is implemented to achieve versioning, data transfer, and upload session. Finally, we discuss the implementation of the metadata manager and provenance manager that supports metadata and provenance graph retrieval. 

\subsection{Credential Server}

   \begin{figure}[!htb]
        \begin{center}
        \includegraphics[scale=0.15]{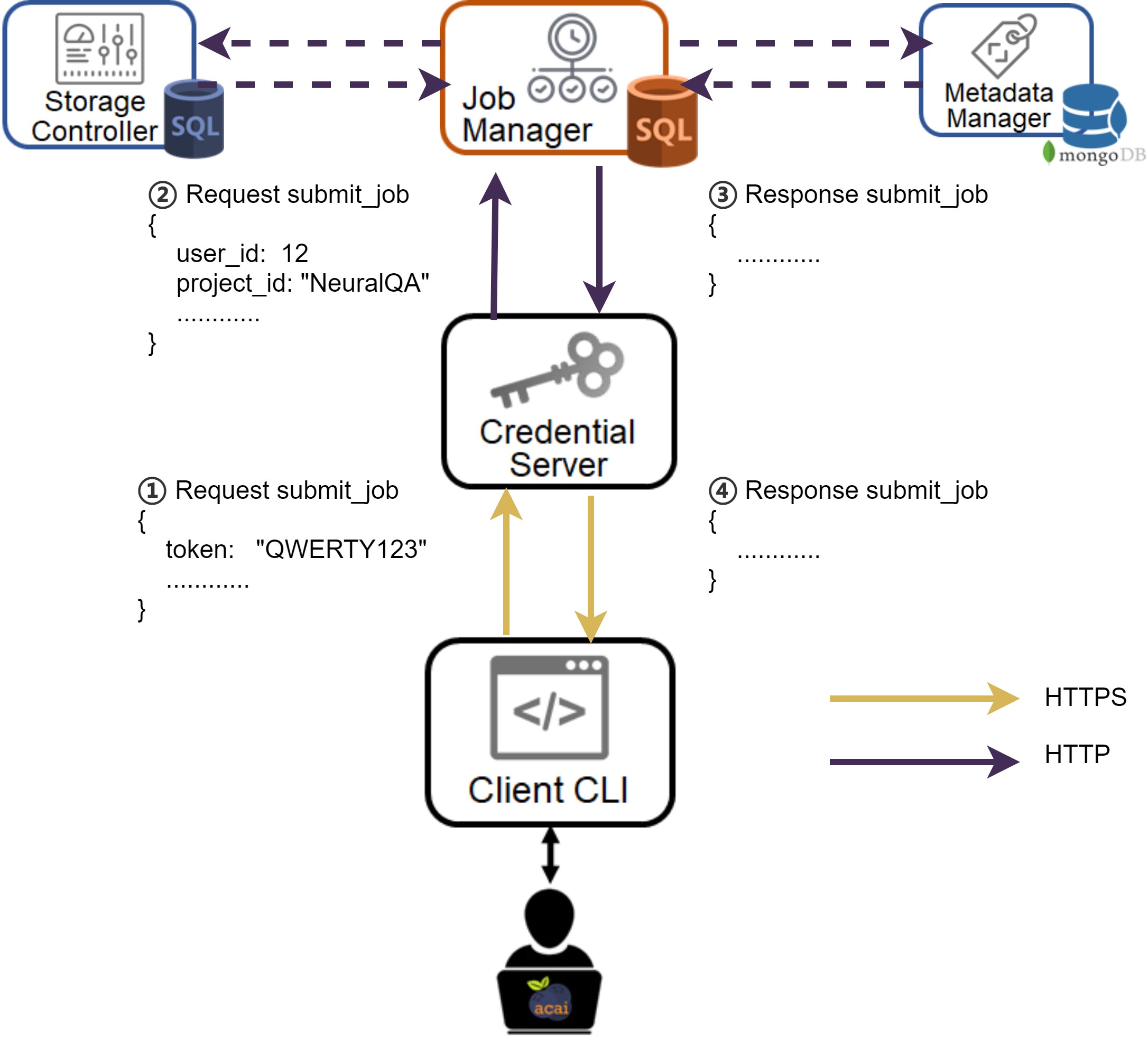}  
        \caption{Example workflow of request authentication and redirect}
        \label{fig:credential}
        \end{center}
    \end{figure}

The credential server is the only endpoint of ACAI that is open to the clients. It authenticates requests from the client and redirects them to corresponding services (see \autoref{fig:credential}).

Requests are authenticated by the user token. The user token is generated randomly upon user creation and should be attached to every request issued by the client. Upon receiving a request, the credential server authenticates the token, and if the authentication succeeds, parses the token into an internal user id and its corresponding project id, and redirects that request. 

HTTPS is used between the platform and external clients in order to guarantee security. An apache reverse proxy server will redirect external requests to the credential server as HTTP requests, and internal communications of the platform use HTTP to boost performance.

\subsection{Execution Engine}
\begin{figure}[!htb]
        \begin{center}
        \includegraphics[scale=0.45]{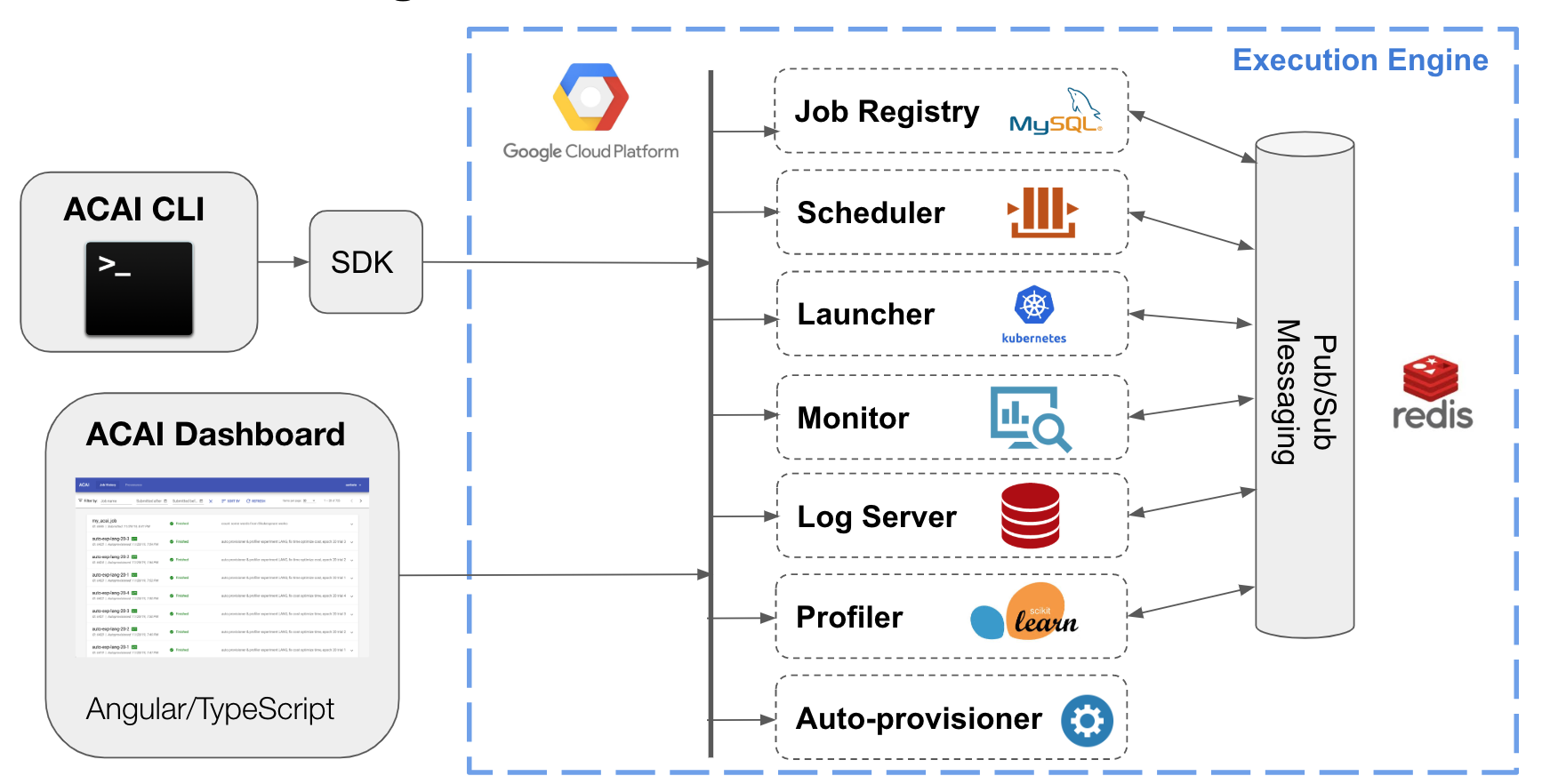}  
        \caption{Microservices in the execution engine}
        \label{fig:exe-arch}
        \end{center}
    \end{figure}

We implement the execution engine as a set of microservices, which includes job registry, job scheduler, job launcher, job monitor, log server, profiler, and auto-provisioner,  as shown in \autoref{fig:exe-arch}. \textbf{Job registry} is a microservice that maintains a repository of all submitted jobs with their respective metadata. \textbf{Job scheduler} maintains a job queue for each user and uses a quota-based, FIFO scheduling strategy to determine when to launch a specific job. \textbf{Job launcher} is responsible for provisioning job containers inside the Kubernetes cluster Engine and watching container status inside the cluster. \textbf{Job monitor} keeps track of runtime information of all submitted jobs. \textbf{Log server} reads logs from job containers and persists them for future references. Profiler trains predictive models for job runtime by exploring a subset of the full configuration space. \textbf{Auto-provisioner} optimizes the runtime or cost of a job based on the predictions made by the profiler. 

The microservices coordinate with one another via RESTful APIs and a Redis event bus that supports pub/sub messaging paradigm (see \autoref{fig:exe-arch}). In the pub/sub messaging paradigm, messages published to a topic by a microservice are immediately received by other microservices that subscribe to the same topic. We designed two primary topics in our system: container status topic and job progress topic. The container status topic is managed by a job launcher who monitors and publishes real-time container status in the Kubernete cluster. The job progress topic is used to publish real-time job progress updates (downloading, running, uploading, etc.).

\subsubsection{Job Execution Flow}

\begin{figure}[!htb]
        \begin{center}
        \includegraphics[scale=0.4]{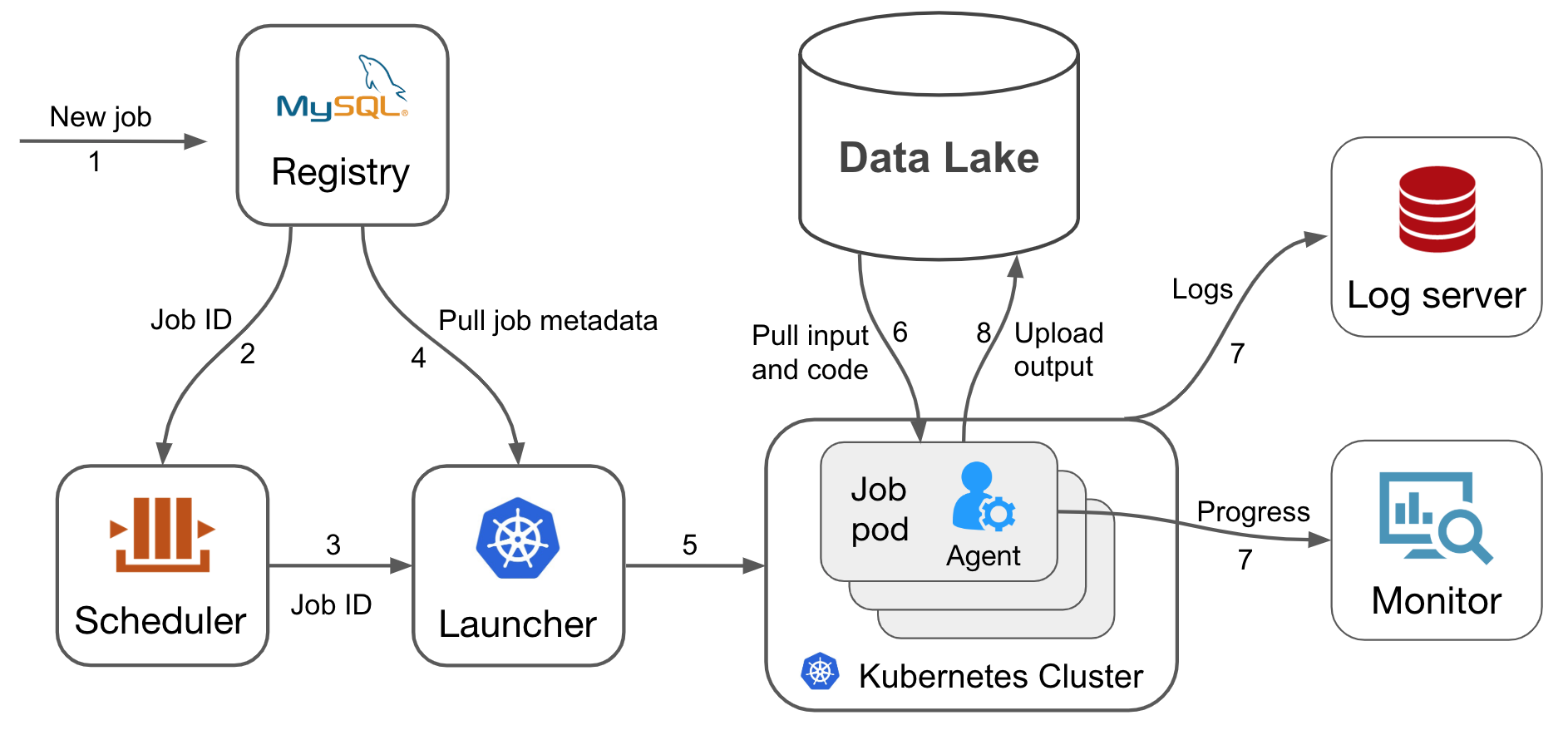}  
        \caption{Job execution flow}
        \label{fig:job-flow}
        \end{center}
    \end{figure}

As shown in \autoref{fig:job-flow}, the job execution flow starts when the job registry receives a new job submission. The job registry first assigns the job with a unique job ID and persists in the job’s metadata in the database. The job registry then calls the job scheduler to enqueue the job ID for launching. When the job scheduler decides to launch a job, it polls the job ID from the job queue and calls the job launcher with the job ID. Job launcher then fetches job metadata from the registry and provisions a new container in the Kubernetes cluster to run the job. 

Each job container has a pre-installed, long-running agent program that manages and supervises job execution in the container. Once the container is launched, the agent starts by downloading code and an input file set from the data lake and executes the job with the command provided by the user. When the job finishes, the agent uploads the output file set to the data lake. During this process, the agent broadcasts job progress to the Redis event bus in real-time, and job monitor subscribes to the topic to receive job status updates. In the meantime, the log server reads logs from the Kubernetes cluster generated by the user’s program, persists logs in its local disk, and sends the logs to the job monitor. The job monitor maintains WebSocket connections with the dashboard to display the latest job status and logs to the user.

When the job finishes, the job scheduler is notified via the message channel and schedules the next job if the user has pending jobs in the job queue. The provenance server will also be notified upon the job’s successful completion and updates the provenance graph with the new job and output file set.

\subsubsection{Job Profiling: Learning to Predict Runtime}
Runtime prediction is a supervised learning problem. When a user requests to profile an ML job, the user supplies a command template that contains hints of arguments of interests where each hint is a set of integers, $\texttt{opts}_i$. An example invocation of profiling and auto-provisioning using the ACAI CLI is as follows, \\

\begin{verbatim}
    $ acai profile --template_name my_template \
        --command_template  'python train.py \
            --epoch {1,2,5} \ 
            --batch-size {256,1024} \
            --learning-rate 0.001'
    $ acai autoprovision --template_name my_template --values 100 256
\end{verbatim}

Suppose the command line contains k hints, then the profiler launches $$\vert  \texttt{cpus} \vert \vert\texttt{mems}\vert \prod_{i=1}^k \vert\texttt{opts}_k\vert $$ profiling jobs to obtain a training set, where $\texttt{cpus}$ and $\texttt{mems}$ is a set of CPU configurations and a set of memory configurations to explore respectively. We set $\texttt{cpus}=\{0.5,1,2\}$ and $\texttt{mems} = \{512,1024,2048\}$ to reduce the exploration space and profiling cost. The profiler waits for 95\% of the profiling jobs to finish before starting training a runtime prediction model to cope with straggler jobs. 

Once the runtime prediction model is trained and ready for serving, the profiler exposes an endpoint for querying the runtime of a command template by supplying an instance resource configuration and command-line template variables. 

\subsubsection{Profiling Model}
\begin{figure}[!htb]
        \begin{center}
        \includegraphics[scale=0.5]{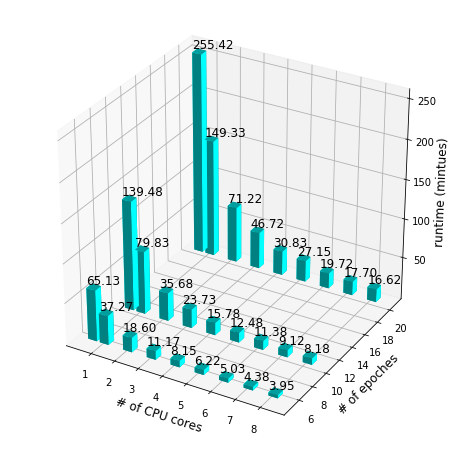}  
        \caption{Relationship between job runtime and CPU and the number of epochs}
        \label{fig:3dbar}
        \end{center}
    \end{figure}

We use a PyTorch deep learning task as an example, and measure the implication on the runtime of both the number of CPUs and the number of epochs. \autoref{fig:3dbar} shows that the runtime is approximately $t_1 \cdot e \cdot c^{-1}$ where $c$ is the number of CPU cores, $e$ is the number of epochs and $t_1$ is the time to train the model for one epoch. Based on the case study, we propose a log-linear model for runtime prediction given resource configurations. Suppose the configuration space has $k$ dimensions, then the runtime is multiplicative in terms of the features in the configuration space,
$$  y = \alpha \prod_{i=0}^k x_i^{\beta_i} $$  
where $x_i$ represents a feature in the configuration space, and y represents the runtime. $\alpha$ and $\beta_i$’s are learnable parameters. We obtain a linear model by taking the log on both sides.
$$  \log y = \log \alpha + \sum_{i = 0}^k \beta_i \log x_i $$
And thus, we train a linear regression model where $\log y$ is the dependent variable and $\log x_i,\cdots,\log x_k$ as explanatory variables. The training set is obtained by the profiler.

\subsubsection{Optimizing Runtime and Cost}
When a user auto-provisions a job that has been profiled, the user specifies the values to use in the command template, along with the maximum runtime or cost that can be tolerated. For auto-provisioning, we support 512MB to 8192 MB memory with a granularity of 256 MB, and 0.5 to 8 vCPUs with a granularity of 0.5 vCPU. Since the configuration space is discrete, the auto-provisioner searches for the optimal vCPU and memory configuration by grid search. Specifically, the auto-provisioner queries the profiler for a predicted runtime for each possible configuration. Based on the predictions from the profiler, the auto-provisioner first filters out configurations that would exceed the user’s runtime or cost constraints. Then it finds the configuration that optimizes runtime or cost, composes a new job using the configuration, and submits the job to the job registry.

\subsection{Cloud Pricing Model}
The execution engine charges users based on job resource configurations and runtime. Users can provision a minimum of 0.5 vCPUs and 512 MB memory, and a maximum of 8 vCPUs and 8192 MB memory for each job. Each vCPU and each MB of memory provisioned for a job is billed separately. The unit price for vCPU is defined as the price per vCPU per hour, and the unit price for memory is defined as the price per GB of memory per hour.

    \begin{figure}[ht]
        \begin{center}
        \includegraphics[scale=0.6]{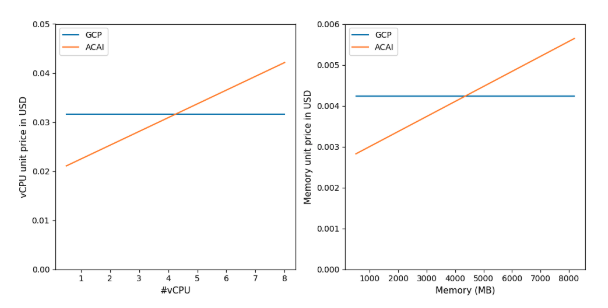}
        \caption{Cloud pricing model}
        \label{fig:pricing}
        \end{center}
    \end{figure}

As a baseline of our pricing model, we use GCP’s N1 machine type on-demand resource pricing in the us-east1 region. We set the unit vCPU price to be 2/3 of GCP’s price for a job with 0.5 vCPUs, and 4/3 of GCP’s price for a job with 8 vCPUs. Similarly, we set the unit memory price to be $\frac{2}{3}$ of GCP’s price for a job with 512 MB memory, and 4/3 of GCP’s price for a job with 8192 MB memory (see \autoref{fig:pricing}). The unit price for both vCPUs and memory increases linearly with the amount of vCPUs and memory provisioned. The reason to adopt variant unit prices is to encourage users to provision jobs with fewer resources due to the added cost of vertical scaling.

\subsection{Data Lake}
In this section, we first introduce the implementation of the versioning of files and file sets using the cloud object storage and MySQL database. Then, we show how data are transferred between the data lake and the clients. We also explained how we support efficient uploading of file batches.

\subsubsection{Versioning}
ACAI stores the vast amount of user data on Amazon S3, with each user file as an S3 object. The actual file hierarchy is stored in a MySQL table. Another table is used to store file sets, and each file set is stored as a row in the table with its content as a list of pairs of a file name and a version number.

Instead of relying on S3 to support file versioning, we implement versioning on top of S3 to avoid cloud vendor lock-in. We use similar ideas to implement versioning for both files and file sets. Essentially, every version is treated as an immutable entity with a row in the database table, and, for files, an object on S3. On file upload or file set creation, a server-side lock is used to guarantee sequential version number assignment.

\subsubsection{Data transfer}
To avoid unnecessary traffic on ACAI servers and also improve data safety, user data does not go through ACAI servers. Instead, to upload or download files, the client first sends a request to the storage server to get the Amazon S3 presigned URLs. It then directly communicate with S3 with the presigned URLs. The storage server subscribes to a preconfigured Amazon SNS so that when an object is uploaded to or downloaded from S3, SNS sends a notification to the storage server.

\subsubsection{Upload session}
Batch file uploading can be tricky because file uploads in a versioning system must guarantee that:

\begin{itemize}
    \item No asynchronous uploads should overwrite each other. In our case, files must be uploaded to different paths on S3.
    \item Asynchronous uploads to the same file name must be sequentially numbered versions.
    \item Failed uploads must not occupy version numbers so that there are no gaps between version numbers. 
\end{itemize}

\begin{figure}[!htb]
        \begin{center}
        \includegraphics[scale=1.0]{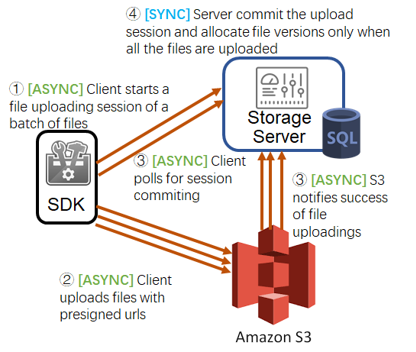}
        \caption{Upload session workflow}
        \label{fig:upload_session}
        \end{center}
    \end{figure}

Therefore, it is necessary to introduce transactional semantics to give those guarantees. A transaction to upload a batch of files is called an upload session. The client sends a request to the storage server with all the file names it wants to upload to start an upload session. The session starts with an initial state named pending. The storage server allocates a unique numerical file id for each file to upload, and uses that id as the S3 uploading destination path. Once receiving the list of presigned URLs from the storage server, the client uploads all the files to S3 and it keeps polling the server until the server confirms that the upload session is committed. Meanwhile, S3 notifies the server of succeeded file uploads. Once all the files in the session are uploaded. The storage server commits the upload session by allocating version numbers to files in the upload session and change the state of the session to committed. The sessions are committed sequentially to avoid version collision. Session states are persisted in the database, so that after the client or server crashes, the states are not lost, and the client is free to either continue the session or abort it. The storage server aborts a session by deleting all the files that are already uploaded to S3 and clean up the session states in the database. \autoref{fig:upload_session} illustrates a simplified workflow of a successful upload session.

\subsection{Metadata and Provenance Server}
Metadata are key-value attributes of files, file sets, and jobs. And the provenance graph is a directed acyclic graph of file sets (node) and actions (edge) where an action is either a job execution or a file set creation. The metadata server and provenance server of ACAI provide access to metadata and the provenance graph. Key-value attributes are retrieved through key matching while the provenance graph is retrieved through semantic queries that use the graph structure. Given these different query mechanisms, we store key-value attributes in a document store, MongoDB\cite{banker2011mongodb}, and provenance graph in a graph database, Neo4j\cite{vukotic2014neo4j} for faster retrieval. The major downside of using two different databases though are (1) increased maintenance cost and (2) increased number of queries when retrieving the metadata of a node or edge in the provenance graph because users have to query both databases. But the performance gain outweighs the downsides. 

We discuss the implementation details of each server in the below sections:
    \subsubsection{Metadata Server}
    The metadata server is hosted on top of a document store, MongoDB. MongoDB is a document-oriented NoSQL database used for high volume data storage. It uses JSON-like documents with schema and allows custom key-value pairs. In addition, it supports data indexing, range query, regular expression query, among other things.
 
    Artifacts of different projects are stored in different tables and each artifact is a document in MongoDB. At file upload, file set creation and job completion, the storage engine and execution engine send a RESTful request to the metadata server with the project id as well as the artifact id and default key-value attributes of an artifact. The metadata server allows adding user-defined key-value attributes. However, to regulate user-customized metadata, we predefine some indexed keys with null values that users may update when appropriate. For example, every job in the system has an attribute “training\_loss”.  For each key of key-value attributes, the metadata server will create an index for a key if it does not exist in the database. This will boost performance for queries but will increase the storage cost.
    
    \subsubsection{Provenance Server}
    The provenance server is hosted on top of a graph database management system, Neo4j. It is an ACID-compliant transactional database with native graph storage and processing. The data model of Neo4j comprises of two main components, Node, and Relationship. Nodes are the entities of a graph and in our case represent file sets in ACAI and Relationships are directed and named connections between two nodes which are job execution and file set creation in ACAI. 

    In the provenance server, each node is uniquely identified by its project id and file set id while each edge is identified by its project id, job id and job type. For each node and relationship in the provenance server, we only store their id while their metadata is kept in the metadata server. The id field is indexed for fast retrieval.  At file set creation and job completion, the storage engine and execution engine will send a RESTful request with input and output file set id as well as the relationship id. The provenance server will insert the nodes and relationships accordingly. The provenance server provides three main APIs to (1) get the whole provenance graph, (2) traverse forward by one edge from a node and (3) traverse backward by one edge from a node. Users interact with the provenance server through the ACAI dashboard where a user can either (1) retrieve the whole graph or (2) search a node with id and then trace forward or backward.

\section{Experiments}
In this section, we measure the effectiveness of resource auto-provisioning and the usability of the ACAI system as a whole. We first conduct a series of auto-provisioning experiments using a program that performs deep learning using the PyTorch framework. We analyze the prediction error of code runtime, and we show the speed-up and cost reduction achieved using the resource configuration recommended by the auto-provisioner. Also, we compare our system to GCP on a hyperparameter-tuning task along the dimension of environment set up, model training, and experiment tracking. Finally, we show qualitative evaluations of our system by interviewing an early user.  

\subsection{Auto-provisioning Experiments}
We test the efficacy of our profiler and auto-provisioner on the MNIST hand-written digit recognition task. We use the program included in PyTorch as one of the official examples. The program trains a multi-layer perceptron using the MNIST dataset. For each epoch, the program streams the training data in batches and updates the model using batch gradient descent. A cross-validation is performed at the end of an epoch. In this set of experiments, we compare the predicted runtime to the true runtime, and we measure the speed up and cost savings using our recommended resource configurations.  

\subsubsection{Runtime Predictions}
\begin{figure}[!htb]
        \begin{center}
        \includegraphics[scale=0.6]{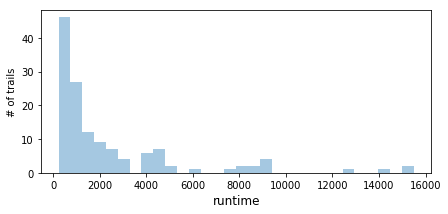}
        \caption{Distribution of the runtime of evaluation trials}
        \label{fig:dist}
        \end{center}
    \end{figure}

We train the profiler using a parameter space spawned by the Cartesian product of $\texttt{epoch}=\{1, 2, 3\}$, $\texttt{cpu}=\{0.5, 1, 2\}$, and $\texttt{memory}=\{512\texttt{MB}, 1024\texttt{MB}, 2048\texttt{MB}\}$, and thus the training data set contains 27 trial runs. During evaluation, we run the Python program using $\texttt{epoch} \in \{5,10,20\}$, $ \texttt{cpu} \in \{0.5, 1, 2, 3, 4, 5, 6, 7, 8\}$, and $\texttt{memory} \in \{512 \texttt{MB}, 1024 \texttt{MB}, 2048 \texttt{MB}, 4096 \texttt{MB}, 8192\texttt{MB}\}$. The evaluation dataset contains 135 trials. \autoref{fig:dist} shows the distribution of the runtime of the evaluation trials.
    
    The average runtime is 2105.71 seconds in the evaluation set. We measure L1 (mean absolute error) and L2 errors (mean squared error) of the predicted runtime (see \autoref{table:pred-error}). We compare our model with averaging the runtime of evaluation trials. The results show that our model explains 98\% of the observed variance in the evaluation trials. 

\begin{table}[ht]
\centering
{\renewcommand{\arraystretch}{1.2}
\begin{tabular}{|l|l|l|}
\hline
\textbf{Model}                            & \textbf{L1 error (seconds)}               & \textbf{L2 error ($\texttt{seconds}^2$) }    \\ \hline
Averaging runtime in eval trials & 2105.71                          & 9375932.21                          \\ \hline
Log linear regression            & \textbf{224.82} &  \textbf{194173.04} \\ \hline
\end{tabular}
}
 \caption{Runtime prediction error}
 \label{table:pred-error}
\end{table}

\begin{figure}[!htb]
        \begin{center}
        \includegraphics[scale=0.5]{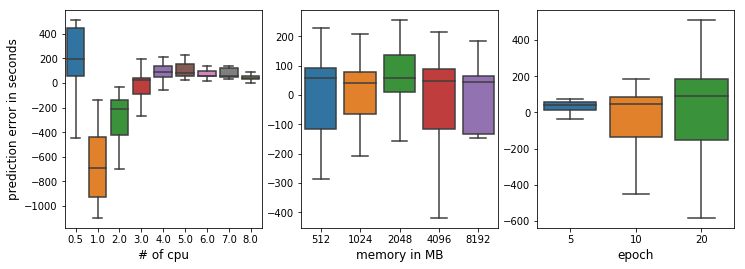}
        \caption{Error with respect to the number of CPU cores, memory amount, and the number of epochs}
        \label{fig:error1}
        \end{center}
    \end{figure}

\autoref{fig:error1} shows the error with respect to the number of CPU cores, memory and the number of epochs. The CPU error plot shows that the error variance is higher when there are fewer CPU cores. One possible explanation is that there are more context switches when there are fewer CPU cores. The context switch overhead has a high variance since it depends on which processes are scheduled before or after the training process. The CPU error plot also shows a non-linearity, which suggests higher-order terms in the relationship between runtime and the number of CPU cores. The error with respect to the number of epochs plot shows that the variance of prediction error increases when the epoch increases from 5 to 20. Caching, I/O buffering, and multi-tenancy contribute to the variability in the runtime of long-running jobs in the cloud environment.

\begin{figure}[!htb]
        \begin{center}
        \includegraphics[scale=0.5]{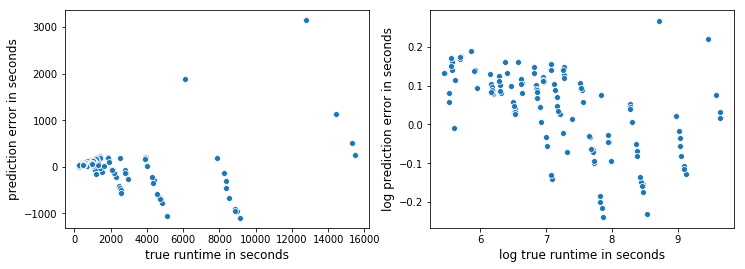}
        \caption{Error with respect to the true runtime}
        \label{fig:error2}
        \end{center}
    \end{figure}

\autoref{fig:error2} shows the prediction error to the true runtime in both linear and log space. In the log space, the log prediction error has uniform variance but shows a non-linear trend. The model underestimates runtime significantly when the true runtime is near 9000 seconds, but it overestimates when the true runtime is greater than 10000 seconds. This nonlinearity is the result of the missing higher order term in the CPU error plot in \autoref{fig:error2}. 

\subsubsection{Optimizing Resource Configurations}
In this set of experiments, we test the effectiveness of the auto-provisioned resource configurations on the MNIST PyTorch task. We run the task for 20 and 50 epochs. The baseline uses the same resource configuration as GCP n1-standard-2 virtual machine type, which includes 2 vCPUs and 7.5GB memory. We run each job three times and record the average runtime for both baselines and auto-provisioned jobs. Cost for each job is calculated using the following formula:
$$ \texttt{Total\_cost} = (\texttt{vCPU\_unit\_cost} \times \texttt{\#vCPU} + \texttt{memory\_unit\_cost} \times \texttt{memory}) \times \texttt{runtime} $$

We first fix the maximum cost to be the same as the cost of the baseline and optimizes for runtime (see \autoref{table:auto-exp-result-1}).

\begin{table}[]
\centering
{\renewcommand{\arraystretch}{1.2}
\begin{tabular}{|l|l|l|l|l|l|l|l|}
\hline
\multirow{2}{*}{Epochs} & \multicolumn{3}{c|}{Baseline}                                                                                                & \multicolumn{3}{c|}{Auto-provisioned}                                                                                           & \multicolumn{1}{c|}{\multirow{2}{*}{Speedup}} \\ \cline{2-7}
                        & Resource                                                & \begin{tabular}[c]{@{}l@{}}Avg.\\ Runtime\end{tabular} & Avg. Cost & Resource                                                   & \begin{tabular}[c]{@{}l@{}}Avg.\\ Runtime\end{tabular} & Avg. Cost & \multicolumn{1}{c|}{}                         \\ \hline
20                      & \begin{tabular}[c]{@{}l@{}}2 vCPUs\\ 7.5GB \end{tabular} & 64.6                                                   & \$0.09765 & \begin{tabular}[c]{@{}l@{}}7.5 vCPU\\ 3584MB \end{tabular} & 16.6                                                   & \$0.08837 & \textbf{1.74x}                                \\ \hline
50                      & \begin{tabular}[c]{@{}l@{}}2 vCPUs\\ 7.5GB\end{tabular} & 162.2                                                  & \$0.24519 & \begin{tabular}[c]{@{}l@{}}8 vCPU\\ 3328MB\end{tabular}   & 37.4                                                   & \$0.21800 & \textbf{1.77x}                                \\ \hline
\end{tabular}
}
\caption{MNIST task: Fix maximum cost and optimize for runtime}
\label{table:auto-exp-result-1}
\end{table}

In this experiment, the auto-provisioner provisions 4 times more vCPUs and halves memory compared to the baseline, even though the cost per vCPU increases as the number of vCPUs increases. But for this task, since the batch size is fixed, runtime is agnostic to memory change so that reducing the memory resource can reduce cost without significantly affecting job runtime. To visualize auto-provisioner’s decision making, we plot the predicted runtime of all resource configurations in the search space for the MNIST 20-epoch task.

\begin{figure}[ht]
    \begin{center}
    \includegraphics[scale=0.5]{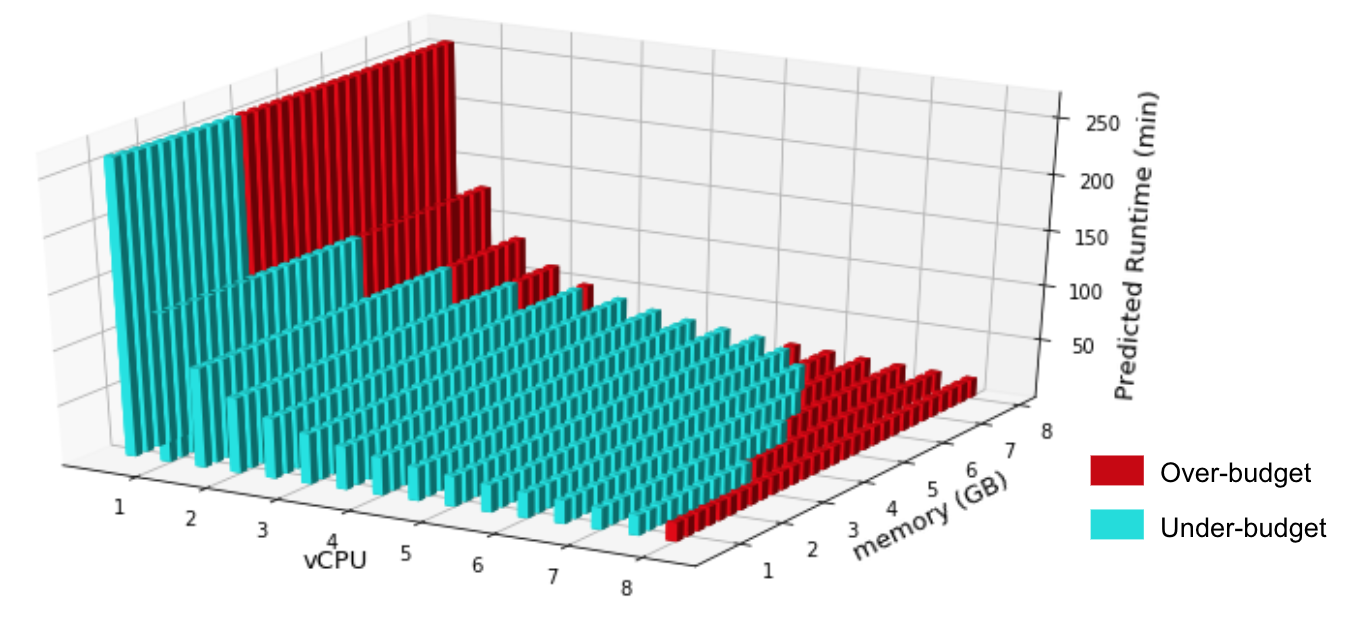}
    \caption{MNIST 20-epoch predicted runtime for every resource configuration}
    \label{fig:mnist-auto}
    \end{center}
\end{figure}

In \autoref{fig:mnist-auto}, resource configurations that have estimated costs higher than the baseline cost \$0.09765 are displayed in red, which are discarded by the auto-provisioner. These configurations are over budget either because the runtime is too long (upper left part), or because the resource unit price is too high (lower right part). Among the under-budget resource configurations, auto-provisioner chooses the one with the lowest estimated runtime. The results show that the auto-provisioner achieves 1.7x speed up without incurring more costs.

We then fix the maximum runtime to be the same as the baseline and optimize for cost (see \autoref{table:auto-exp-result-2}). 

\begin{table}[]
\centering
{\renewcommand{\arraystretch}{1.2}
\begin{tabular}{|l|l|l|l|l|l|l|l|}
\hline
\multirow{2}{*}{Epochs} & \multicolumn{3}{c|}{Baseline}                                                                                                & \multicolumn{3}{c|}{Auto-provisioned}                                                                                         & \multicolumn{1}{c|}{\multirow{2}{*}{Speedup}} \\ \cline{2-7}
                        & Resource                                                & \begin{tabular}[c]{@{}l@{}}Avg.\\ Runtime\end{tabular} & Avg. Cost & Resource                                                 & \begin{tabular}[c]{@{}l@{}}Avg.\\ Runtime\end{tabular} & Avg. Cost & \multicolumn{1}{c|}{}                         \\ \hline
20                      & \begin{tabular}[c]{@{}l@{}}2 vCPUs\\ 7.5GB\end{tabular} & 64.6                                                   & \$0.09765 & \begin{tabular}[c]{@{}l@{}}2.5 vCPU\\ 512MB\end{tabular} & 52.6                                                   & \$0.05975 & \textbf{38.8\%}                               \\ \hline
50                      & \begin{tabular}[c]{@{}l@{}}2 vCPUs\\ 7.5GB\end{tabular} & 162.2                                                  & \$0.24519 & \begin{tabular}[c]{@{}l@{}}2.5 vCPU\\ 512MB\end{tabular} & 140.4                                                  & \$0.15949 & \textbf{35.0\%}                               \\ \hline
\end{tabular}
}
\caption{MNIST task: Fix maximum time and optimize for cost}
\label{table:auto-exp-result-2}
\end{table}

To reduce costs as much as possible, the auto-provisioner provisions the least amount of memory (512MB) possible. It also provisions 0.5 more vCPUs to compensate for the reduction of memory in order to meet the maximum run time constraint. \autoref{table:auto-exp-result-2} shows the auto-provisioner reduce cost by over 30\% while keeping the job run time less than the baseline. Compared to the previous experiment which optimizes for run time, the provisioner is much more conservative in resource allocation in this experiment, which makes the job run longer but cost less.

    \subsection{Usability Study}
    The objective of ACAI is to accelerate experiment workflows for machine learning scientists. To evaluate the achievement of ACAI, we design and conduct a usability study to show that ACAI:
    \begin{enumerate}
        \item Reduces the experiment set-up time.
        \item Reduces experiment bookkeeping time.
        \item Reduces the overall time and cost of experiments.
    \end{enumerate}
    In this usability study, we compare the time and cost of completing a hyperparameter tuning task for a typical machine learning model manually using GCP virtual machines against using ACAI SDK. Hyperparameter tuning is a process where a practitioner trains a set of models with different values of a set of hyperparameter using the training dataset and then evaluates the trained models with the testing dataset. The practitioner then, based on model performance, decides the best set of hyperparameters. 

    The usability study comprises of two rounds of controlled experiments where the control performs the task on GCP and the treatment uses ACAI SDK. After the first controlled experiment, the control and treatment will switch roles and tune a different model. This is to ensure that the result from this study is not affected by participants’ proficiency in one particular development setting. Given the limits in time and resources, the control and treatment group each has one tester for this study.

    In each controlled experiment, both groups are presented with a problem, a machine learning model and a set of hyperparameters to tune. The details of problems and machine learning models are presented in \autoref{task_definitions}. We fix the search space for the set of hyperparameters and therefore the number of training and evaluation jobs is also fixed. The computation resource is fixed at 8 CPU, 64GB memory and 2 Tesla P100 GPUs. Both groups are provided with a training dataset, evaluation dataset, training script and evaluation script hosted on AWS S3.  The control group is free to develop its own job batching and scheduling script while the treatment group uses ACAI SDK. The time spent on each stage of the workflow (as categorized in \autoref{table:experiment_time}) is recorded:
    \begin{table}[ht]
      \centering
      {\renewcommand{\arraystretch}{1.2}
      \begin{tabular}{|c|c|}
        \hline \textbf{Category}   & \textbf{Items} \\ 
        \hline Experiment set-up   & \begin{tabular}[c]{@{}c@{}}1. Resource deployment\\ 2. Code development\end{tabular} \\ 
        \hline Experiment Tracking & Bookkeeping experiment results \\ 
        \hline Total Time & The total time spent on the experiment  \\ 
        \hline
        \end{tabular}
        }
        \caption{Time Categories}
      \label{table:experiment_time}
    \end{table}\\
    The cost of experiments is calculated by: 
   $$ \texttt{Cost} = \texttt{amount\_of\_resources} \times \texttt{duration\_deployed} \times \texttt{unit\_price\_listed} $$
   
   \subsubsection{Result}
   We present the result of the two controlled experiments in \autoref{table:conroll_1} and \autoref{table:conroll_2}. We show the time spent on different categories by both the control and treatment group as well as the money spent by both groups. We also calculate the improvement in time and cost. The result shows that, in both controlled experiments, using the ACAI SDK results in a reduction in time in every category we recorded. Code development is easier with the ACAI SDK as reflected in more than a 20\% savings in code development time. Since ACAI takes care of resource deployment, there is no time spent on resource deployment. We also observe a notable saving in experiment tracking time of more than 40\%\footnote{The saving in experiment tracking time is much more significant in the second controlled experiment and this may be due to (1) the increased amount of jobs and (2) in the second controlled experiment, both groups are asked to try all possible combinations of hyperparameter and in the first controlled experiment, both groups are required to find the best value for a hyperparameter before proceeding to the next one which introduces extra bookkeeping time even using ACAI SDK.}. The reduction in total experiment time also translates to more than 2\% of cost savings. 
   \begin{table}[ht]
      \centering
      {\renewcommand{\arraystretch}{1.2}
        \begin{tabular}{|c|c|c|c|c|}
        \hline
        \multicolumn{5}{|c|}{\textbf{\begin{tabular}[c]{@{}c@{}}Experiment Round 1 (MLP)\\ Total Number of Jobs: 16\end{tabular}}}  \\ \hline
        \multicolumn{2}{|c|}{}                                                                                                             & \textbf{\begin{tabular}[c]{@{}c@{}}Control\\ (GCP)\end{tabular}} & \textbf{\begin{tabular}[c]{@{}c@{}}Treatment\\ (ACAI SDK)\end{tabular}} & \textbf{Improvement} \\ \hline
        \multirow{2}{*}{\textbf{Experiment Setup Time}} & \textbf{\begin{tabular}[c]{@{}c@{}}Code Development\\ {[}min{]}\end{tabular}}    & 21.47  & 16.65 & 22\%  \\ \cline{2-5} 
        & \textbf{\begin{tabular}[c]{@{}c@{}}Resource Deployment\\ {[}min{]}\end{tabular}} & 14.37                                                            & 0                                                                       & -                    \\ \hline
        \multicolumn{2}{|c|}{\textbf{\begin{tabular}[c]{@{}c@{}}Experiment Tracking Time \\ {[}min{]}\end{tabular}}}                       & 8.52                                                             & 5.07                                                                    & 40\%                 \\ \hline
        \multicolumn{2}{|c|}{\textbf{\begin{tabular}[c]{@{}c@{}}Total Time\\ {[}min{]}\end{tabular}}}                                      & 188.77                                                           & 148.03                                                                  & 21\%                 \\ \hline
        \multicolumn{2}{|c|}{\textbf{Total Cost}}                                                                                          & \$ 4.666                                                         & \$ 4.502                                                                & 2\%                  \\ \hline
        \end{tabular}
        }
      \caption{Result of $1^{st}$ Controlled Experiment}
      \label{table:conroll_1}
    \end{table}\\
    
    \begin{table}[ht]
      \centering
      {\renewcommand{\arraystretch}{1.2}
        \begin{tabular}{|c|c|c|c|c|}
        \hline
        \multicolumn{5}{|c|}{\textbf{\begin{tabular}[c]{@{}c@{}}Experiment Round 2 (Xgboost)\\ Total Number of Jobs: 72\end{tabular}}}                                                                                                                                                                         \\ \hline
        \multicolumn{2}{|c|}{}                                                                                                             & \textbf{\begin{tabular}[c]{@{}c@{}}Control\\ (GCP)\end{tabular}} & \textbf{\begin{tabular}[c]{@{}c@{}}Treatment\\ (ACAI SDK)\end{tabular}} & \textbf{Improvement} \\ \hline
        \multirow{2}{*}{\textbf{Experiment Setup Time}} & \textbf{\begin{tabular}[c]{@{}c@{}}Code Development\\ {[}min{]}\end{tabular}}    & 4.75                                                             & 2.23                                                                    & 44\%                 \\ \cline{2-5} 
                                                        & \textbf{\begin{tabular}[c]{@{}c@{}}Resource Deployment\\ {[}min{]}\end{tabular}} & 7.43                                                             & 0                                                                       & -                    \\ \hline
        \multicolumn{2}{|c|}{\textbf{\begin{tabular}[c]{@{}c@{}}Experiment Tracking Time \\ {[}min{]}\end{tabular}}}                       & 12.6                                                             & 1.07                                                                    & 87\%                 \\ \hline
        \multicolumn{2}{|c|}{\textbf{\begin{tabular}[c]{@{}c@{}}Total Time\\ {[}min{]}\end{tabular}}}                                      & 90.62                                                            & 62.58                                                                   & 20\%                 \\ \hline
        \multicolumn{2}{|c|}{\textbf{Total Cost}}                                                                                          & \$ 0.272                                                         & \$ 0.242                                                                & 11\%                 \\ \hline
        \end{tabular}
        }
      \caption{Result of $2^{nd}$ Controlled Experiment}
      \label{table:conroll_2}
    \end{table}
\subsection{User Feedback}
We also collected feedback from a machine learning scientist whom we collected user requirements from in the early stage of the project. The feedback is summarized in  \autoref{table:painpoints-requirements-feedback}, which includes the pain points we initially collected from the user, the requirements suggested by the user and the feedback we collect from the user

\begin{table}[]
\centering
{\renewcommand{\arraystretch}{1.2}
\begin{tabular}{|p{4cm}|p{4cm}|p{4cm}|}
\hline
\multicolumn{1}{|c|}{User Pain Points}                                                                                                 & \multicolumn{1}{c|}{User Requirements}                                                       & \multicolumn{1}{c|}{User Feedback}                                                                                                                                                     \\ \hline
User manually maintains an experiment log. This log can grow unorganized with time and more experiments                               & An automatic and persistent experiment tracking mechanism                                    & Experiments conducted on ACAI are automatically tracked and scientists can add descriptions through tagging. This replaces manual log effectively and efficiently                      \\ \hline
It is tedious and time-consuming to attach context information (e.g. hyperparameter \& model changes) to each corresponding experiment & An efficient or even automatic way to tag experiments and files                              & The log parser of ACAI provides an efficient way to tag an experiment as it progresses. The ability to add/modify tag on the provenance page of the Dashboard is also very helpful     \\ \hline
Evaluating experiments on manually maintained experiment logs is time-consuming and inefficient                                        & The ability to search, filter and categorize experiments                                     & The metadata server of ACAI provides a search interface that allows one to perform various kinds of queries into experiment histories. It will be helpful for scientists to evaluate and reflect on experiments                                                                                    \\ \hline
It is challenging to revert back to a specific version of a model that produces certain experiment result, as changes build up         & The ability to retrieve a specific version of model that corresponds to an experiment result
 & ACAI metadata and provenance server allows one to traverse through experiments and identify output model which can then be retrieved from data lake. This resolves the said pain point \\ \hline
User is unsure how to tune computation resources to meet a deadline or keep cost low                                                   & A mechanism to deploy resources according to schedule and budget                             & The profiler and auto-provisioner of ACAI has the potential to bring down the cost for a research project and help on-time delivery of research results                                \\ \hline
\end{tabular}
}
\caption{User pain points, requirements and feedback}
\label{table:painpoints-requirements-feedback}
\end{table}

The user also commented that ACAI is ideal for experiments where the model is relatively mature and needs (1) hyperparameter tuning and (2) incremental changes to model. It is less suitable for developing a model from scratch, which requires iterative debugging with tools like Jupyter Notebook. 

Based on the results from the usability study and the feedback collected from our initial user, we believe ACAI can boost the efficiency of machine learning scientists and provides a better tool to manage machine learning projects.


\section{Discussion and Conclusion}
We identified that currently the machine learning community is lacking an end-to-end solution that efficiently manages data and models, orchestrates cloud resources and tracks experiments and provenance information. We developed and introduced ACAI, an end-to-end cloud based machine learning platform that provides a data lake and an execution engine. The data lake persists versioned files, file sets and their metadata and the execution engine executes ML jobs on the cloud with auto-provisioning, logging and provenance tracking. 

In evaluations of our system, we demonstrated that the auto-provisioner of ACAI is able to achieve a 1.3x speed up in an MNIST hand-digit recognition task under a budget constraint. Through a usability study, we find that ACAI reduces the time required for a hyperparameter tuning task by 20\% and the cost by 2\%. The ACAI system has already been deployed for a group of research students at Carnegie Mellon University to complete their capstone project. The feedback collected from this group of students and a machine learning scientist who we collected user requirements from earlier in the project suggests that ACAI:

\begin{itemize}
    \item provides an effective and efficient solution for managing and tracking datasets, models and experiments for a machine learning project through tagging artifacts and provenance retrieval;
    \item is helpful in comparing experiments and models;
    \item is efficient in scheduling and launching batch jobs;
    \item improves the efficiency of machine learning scientists by enabling them to reexamine experiments results through provenance server and retrieve a certain version of models through the metadata server and storage service;
\end{itemize}

\section{Future Work}
ACAI is currently a minimum viable product. We have implemented high priority features to have initial feedback form users, but we want to add many more features to make ML scientists more productive. ACAI is also a research project. Our auto-provisioning experiments show a proof of concept of using ML in predicting job behavior. It remains future work for us to test our model in more challenging settings. Also, we have conducted a quick usability study on the ACAI system, but the experiment lacks robustness due to the small sample size. In this section, we discuss future work for the data lake and execution engine subsystems, as well as system evaluations.

\subsection{Data Lake}
We identify the following features that can be added to better support user requirement and boost platform performance:

\subsubsection{Fine-grained access control}
Every request made to the ACAI backend servers is on behalf of a certain user. The credential server performs authentication before redirecting. However, ACAI does not have file or file set level access control. Users can access any resources within the same project. It would be beneficial to support POSIX access control. For example, for every file and file set, ACAI records its read/write permissions for different users and user groups, and does permission checks on every request.

\subsubsection{Inter-job data caching}
ACAI cloud storage does not provide caching. Every job execution starts from downloading data from S3 and ends with uploading data back to S3. As a result, data cannot be directly reused between job executions. It might be beneficial to add a cloud file system acting as a cache that can be mounted and unmounted to share data between job executions. It can be tricky to share cache among arbitrary jobs, as files may have different versions, but it should be fine to share cache between consecutive jobs where the successive job takes in the entire output file set of the precedent job as the input file set.

\subsubsection{Data cleaning and workflow replay}
ACAI persists all the intermediate and output data for the user, but in the long run, data may need to be cleaned to reduce cloud storage cost, either explicitly by users or heuristically by the system. Users should be able to safely delete files that are not part of any job execution. 
Intermediate data that are generated by job executions can also be deleted and ACAI should be able to replay workflows to generate the data again. The run time and cost of history jobs can be good heuristics for users to make decisions on whether to keep or delete intermediate data.

\subsection{Execution Engine}
ACAI only supports single-node ML jobs. But big data ML jobs often involve distributed computation. For example, Apache Spark is a distributed computing framework which scales computation to multiple nodes using the Resilient Distributed Dataset (RDD) abstraction \cite{zaharia2016apache}. Tensorflow supports distributed training across multiple machines by sharding or replicating the computation graph \cite{abadi2016tensorflow}. Supporting distributed computing frameworks is one of the most desired features we want to implement in ACAI execution engine. Also, ML scientists often adopts interactive programming tools such as Jupyter notebooks in debugging ML programs or exploratory data analysis. While ACAI focuses on running ML jobs in batch mode at the moment, we want to support Jupyter notebooks in the future. 

In the report, we show that our job runtime prediction model explains 98\% of observed variance in the runtime of the MNIST pytorch job. But error analysis shows that there is non-linearity between prediction error and the number of CPU cores. This suggests the need for future improvement in the model selection. Also, predicting the runtime of the MNIST example is not very challenging, since it has three input dimensions, the number of epochs, CPU cores and memory. It remains future work for us to test the effectiveness of our runtime prediction model on more input features in typical ML jobs, such as the number of trees when training Random Forest, or the number of layers in a deep neural network. One direction of improving the runtime prediction model is adding more features. Job runtime metrics, such as the CPU utilization and disk IO, are easily accessible within the execution engine. These are time series features, but it is non-trivial to add them to our existing model. The runtime prediction can also utilize data across different jobs using techniques such as transfer learning. This can significantly reduces the cost to profile a new job. 

Job runtime prediction becomes more practical in the context of distributed training since cluster tuning is hard. For example, we can experiment with predicting Spark job runtime conditioned on the number of nodes, and CPU and memory configuration of the master node and work nodes. In this case, our feature space comprises Spark configurations and job specific arguments such as the number of epochs. We still model runtime prediction as an instance of supervised learning, but the feature engineering and model selection process can be more involved. If this works, then we show that  ML is practical in cluster tuning. 

Finally, we want to support ML pipelines for experiment reproduction as well as data and model updates. ACAI stores the dependency of jobs and file sets in the provenance graph, but the provenance graph is read only. There are many cases where users want to rerun a subgraph. For example, if an upstream file set in a subgraph updates, then users might want to update downstream models by re-running all jobs in the subgraph. Another example is that users want to reproduce experiment results using a different dataset but with the same code. We define an ML pipeline is a collection of dependent jobs that can be scheduled by the execution engine as a single entity. As future work, we want to support representing and scheduling an ML pipeline in the execution engine.

\subsection{System Evaluations}
We want to evaluate the ACAI system in a more comprehensive way. In the usability study discussed in this report, constrained by the time and cost, the size of control and treatment is 1 and we can only record the savings in time and cost. We are unable to gauge whether ACAI improves the efficiency of machine learning scientists and helps generate better models. In the future, it would be important to design a controlled experiment with larger control and treatment group size and evaluate the changes brought by ACAI in more dimensions. We may ask half of the students in a deep learning class to use ACAI to complete their homework and the other to complete their homework manually on the cloud. We may record the performance of students and the time and money spent by each student on this homework. This will provide better insight into how ACAI helps a machine learning scientists explore different models and generate better results.  

\newpage

\section{Appendix}
    \subsection{Usability Study Task Definitions} \label{task_definitions}
        \subsubsection{ Frame-level Classification of Speech with Multi-layer Perceptrons (MLP) }
        
        This problem is taken from a deep learning course at Carnegie Mellon University\cite{cmudl}. In frame-level classification of speech, we try to identify the phoneme state label for each frame of a speech recordings. The training data comprises 13,000 speech recordings from Wall Street Journal (raw mel-spectrogram frames) and frame-level phoneme state labels. The testing dataset is a set of speech recordings to be classified. The evaluation metric is the prediction accuracy of the model. The participants are asked to search among the below hyperparameters for a multi-layer perceptron neural network:
        
        \begin{table}[ht]
            \centering
            {\renewcommand{\arraystretch}{1.2}
            \begin{tabular}{|c|c|}
            \hline
            \textbf{Hyperparameter} & \textbf{Values} \\ \hline
            Number of layers & {[}5, 7, 9{]}   \\ \hline
            Number of neighboring frames used in prediction & {[}5, 10, 15{]} \\ \hline
            Batch Normalization Layers  & {[}Yes, No{]}   \\ \hline
            Dropout Layers & {[}Yes, No{]}   \\ \hline
            \end{tabular}
            \caption{Hyperparameter Search Space, MLP}
            \label{table:hyper_mlp}
            }
        \end{table}
        
        \subsubsection{ Porto Seguro’s Safe Driver Prediction with XGBoost \cite{kaggle}}
        This problem is taken from \url{Kaggle.com}, a popular machine learning competitions website. In this problem, one will predict the probability that an auto insurance policy holder files a claim given a set of features. We downloaded the training data and split it into a training set (70\%) and a test set (30\%) and we referenced an XGBoost training program from \url{Kaggle.com}. The training set has 535,000 records and the test set has 59,500. The evaluation metric is the normalized Gini Coefficient. The participants are asked to search among the below hyperparameters for a XGboost Classifier:

        \begin{table}[ht]
            \centering
            {\renewcommand{\arraystretch}{1.2}
            \begin{tabular}{|c|c|}
            \hline
            \textbf{Hyperparameter} & \textbf{Values} \\ \hline
            Maximum tree depth for base learners & {[}2, 6, 10{]}   \\ \hline
            Number of trees to fit & {[}200, 400, 600{]} \\ \hline
            Subsample Ratio of the training instance  & {[}0.8, 1{]}   \\ \hline
            Dropout Layers & {[}gbtree, gblinear{]}   \\ \hline
            \end{tabular}
            \caption{Hyperparameter Search Space, Xgboost}
            \label{table:hyper_xgboost}
            }
        \end{table}

\newpage
\bibliographystyle{unsrt}
\bibliography{main.bib}

\end{document}